\def\ARXIV{}
\documentclass{article}
\usepackage[T1]{fontenc}
\usepackage{iclr2027_conference,times}
\usepackage{amsmath,amsfonts,bm}

\def\eqref#1{equation~\ref{#1}}
\def\1{\bm{1}}

\DeclareMathAlphabet{\mathsfit}{\encodingdefault}{\sfdefault}{m}{sl}
\SetMathAlphabet{\mathsfit}{bold}{\encodingdefault}{\sfdefault}{bx}{n}

\usepackage{graphicx}
\usepackage{booktabs}
\usepackage{longtable}
\usepackage{array}
\usepackage{xcolor}
\usepackage{newunicodechar}
\newunicodechar{’}{'}
\usepackage{hyperref}
\usepackage{url}
\usepackage{xurl}
\usepackage{listings}
\usepackage{textcomp}

\newcommand{\todo}[1]{{\color{blue}\textbf{TODO:} #1}}

\graphicspath{{figures/}}
\title{Coding LLMs Can Be Harnessed to Learn \\
Interpretable Models from Data}
\ifdefined\ARXIV
  \iclrfinalcopy
\author{Aditya Kumar$^{1}$, Cassandra A. Cohen$^{2}$, William W. Cohen$^{1}$ \\
$^{1}$Machine Learning Department, Carnegie Mellon University \\
$^{2}$Language Technologies Institute, Carnegie Mellon University \\
\texttt{adityaku@cs.cmu.edu}, \texttt{ccohen2@andrew.cmu.edu},
\texttt{wcohen@cmu.edu}
}

\else
  \author{Anonymous authors}
\fi
\begin{document}
\maketitle
\ifdefined\ARXIV\lhead{Preprint}\fi
\begin{abstract}
Since compact, interpretable classifiers correspond to simple programs, it is natural to ask
if LLM-based coding agents can be used to learn interpretable models from data.  In this paper we present a simple harness that repeatedly prompts a coding LLM to produce the next rule in a rule set, given a sample of labeled examples. We show that this approach is competitive with, and often superior to, state-of-the-art rule-learning methods on tabular data, especially with small training sets, and wins by large margins on text, including against another LLM-based rule learner; that this approach outperforms standard harnesses, and %that it often outperforms bespoke discrete algorithms for producing rules, especially when coupled with appropriate tools. We
that the approach immediately leads to new kinds of interpretable models, since coding agents are not restricted to search a highly restricted hypothesis space.  In addition to the advances made in learning interpretable models, these results are of interest because they introduce a new class of coding benchmarks---construction of interpretable classifiers---which includes quantitative measurements of output quality (i.e., test-set error) and existing mature ``specialist coding algorithms'' (i.e., existing learning methods).
\end{abstract}
% Keep the shorter figure from floating above the title on the first page.
\suppressfloats[t]
\begin{figure}[t]
\centering
\includegraphics[width=\linewidth]{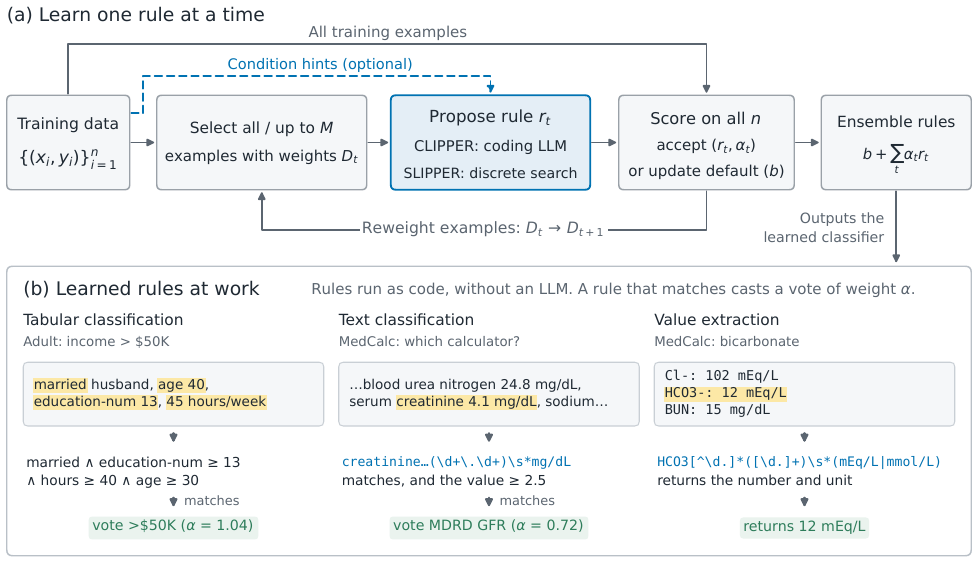}
\caption{Learning a rule classifier with CLIPPER. (a) CLIPPER replaces SLIPPER's
discrete rule search with a coding LLM. The harness iteratively samples the training set using the current distribution; sends the sample to the LLM, prompting it to produce a rule; scores the proposed rule on the full train set; either accepts a rule or updates the default; and reweights
the train examples for the next round. %Optional hints provide full-training statistics.
% OLD: (b) A complete classifier learned on Adult. Rules 1 and 4 are activated by the
% OLD: shown input, giving score $+0.480$ and predicting income above
% OLD: \$50K. Green weights are positive; red weights are negative. Faint bars mark
% OLD: inactive rules.
(b) Summary of three actual CLIPPER-learned rules: attribute
tests correlated with high income; a regex that finds a high creatinine value, correlated with the MDRD GFR calculator; and a regex that extracts a bicarbonate value and corresponding units. %Note excerpts are verbatim from MedCalc test notes; 
Yellow highlights mark relevant parts of the input; 
blue text is copied from a learned rule. % and ``\ldots'' marks
%omitted text. 
%A rule that matches adds its learned weight to the score.
%``Married'' abbreviates \texttt{Married-civ-spouse}, and ``hours'' is hours
%per week.
Note that no LLM is called at inference.
% OLD: ``Married'' abbreviates
% OLD: \texttt{Married-civ-spouse}, ``husband'' is \texttt{relationship = Husband},
% OLD: and ``hours'' is hours per week.
}
\label{fig:tabular}
\end{figure}

% The complete introduction source is immutable, including its old related-
% work stub. Stop reading at that heading and load the new related work below.
\begingroup
\let\OriginalPaperSection\section
\def\section#1{%
  \def\InputSectionTitle{#1}%
  \def\RelatedWorkTitle{Related work}%
  \ifx\InputSectionTitle\RelatedWorkTitle
     
  \else
    \OriginalPaperSection{#1}%
  \fi}
\section{Introduction}
\label{sec:introduction}

%\subsection{Motivation}

Interpretable symbolic models (e.g., rules and decision trees) are a
useful alternative to black-box models in some practically important
settings; however, learning interpretable classifiers remains a
challenging task \citep{rudineuclid2022}.  Since learned interpretable
models correspond to short programs, the development of
LLM-based coding agents raises a natural question: \textbf{how well
  can LLM-based coders produce accurate interpretable symbolic
  models?}  

Past work  \citep{singh2023augimodels,han2024featllm,abhyankar2025llmfe,
zhang2025llmlasso,jia2026tabdistill} shows that LLMs can use their background knowledge to generate \emph{features} for symbolic learners: e.g., FeatLLM uses ensembles of prompted LLMs to generate hundreds of binary features predicting each class, which are then used by a downstream classifier learner.  Much past work on LLM-based feature generation focuses on few-shot learning settings, like FeatLLM, or NLP problems, like Aug-imodels \citep{singh2023augimodels}.  We consider here the more difficult task of \textbf{building complete, fully executable, interpretable classifiers with LLMs}, including tabular classifiers in many-shot settings. There is little prior work on this task, although there has been work on a zero-shot variant of the task \citep{carrasco2025zeroshotdecisiontreeconstruction} and at least one concurrent work \citep{kovacs2026rulechef}: see Section~\ref{sec:related}.

Of course, practitioners often do use coding LLMs to build (or help build)
interpretable models: e.g., in investigating the recent
OpenAI/HuggingFace hacking incident, safety researchers ``iterated
with AI agents to design simple regexes to classify messages''
\citep{metr-2026-openai-hugging-face-incident-investigation}. Of
course, the same AI agents could have been used to directly classify
messages instead, but this was undesirable---elsewhere in the same
report LLM agents were observed to be biased, e.g., to ``present an
overly charitable picture of agents’ reasoning and deceptive
behaviors''
\citep{metr-2026-openai-hugging-face-incident-investigation}. Although we will compare where possible to interpretable models jointly built by humans and LLMs, we focus here on \textbf{fully autonomous systems for building interpretable classifiers from data}.

LLMs have obvious practical advantages over traditional symbolic learners for generating classifiers, just as they do for feature generation: they can exploit background knowledge, and they can use programming language constructs that do not lend themselves to discrete search (like regular expressions).
In addition to the practical value of improving the performance of interpretable learning methods, another motivation for this work is that \textbf{writing predictive rules is different from existing programming benchmarks} like bug-fixes or competitive-programming puzzles.  Test-set error provides a quantitative measure of quality, but not an indication of when a classifier is ``finished''; qualitatively, many LLMs have a tendency to ``overthink'' on this task, generating long, expensive, and unproductive outputs.  There is also a \textbf{strong existing baseline}, since mature
symbolic learners are competent ``specialist coders'' for this task. 

%\subsection{Overview}

\textbf{Our approach is a conservative modification of a classic rule-learning method}.  A harness implements the learner's outer loop: in each iteration, it reweights the training data to focus on examples that are not explained by the accumulated rules, and then submits a sample of them to an LLM agent, prompting it to construct one rule based on the sample.  Specifically, the harness follows the outer loop of SLIPPER \citep{cohen1999slipper}, which uses a
confidence-rated boosting procedure.  This enables a direct comparison of LLM coding agents to SLIPPER's discrete rule search\footnote{Which is based on the search used in RIPPER \citep{cohen1995ripper}.}  to construct
a rule as its inner loop.

We make the following contributions in the paper.

\textbf{A new rule learner.} Experiments with 12 interpretable models on a suite of eight tabular datasets often studied in the interpretability literature
\citep{kruschel2026challenging} show that SLIPPER is a strong baseline (see Figure~\ref{fig:all-baselines}).  Building on this we propose an LLM-based rule learner called CLIPPER (Coding
LLMs used for Iterative Prompting to Produce Error Reduction). 
% OLD: CLIPPER also performs extremely well experimentally on tabular datasets.
On tabular datasets, CLIPPER outperforms existing rule learners with smaller training sets, and is statistically indistinguishable from the strongest baseline given appropriate tools and ``hints'' (relevant statistics that are added to the data sample in the prompt).
{On text datasets, CLIPPER 
% OLD:    performs even better relative to similar baselines, and 
    wins by large margins over similar baselines, including RuleChef \citep{kovacs2026rulechef}, another LLM-based rule learner.  CLIPPER can even be
  extended to use regular expressions that perform extraction, and in
  a comparative study, CLIPPER (using Gemini 3.1 Flash-Lite) found extractors comparable to models constructed
   by  working interactively with a frontier coding agent (Fable 5.1).

\textbf{Analysis and technical innovations}.  Controlled comparisons show that CLIPPER's coding agent consistently outperforms SLIPPER's discrete search given the same information, and additional experiments support the claim that \emph{background knowledge} is the reason.  We also show that the same harness works comparably for multiple model families.  We compare to general-purpose coding harnesses, and show that the CLIPPER harness outperforms them on our test suite.

%The main technical hurdle in using coding LLMs as rule learners is that
%on tabular datasets, \textbf{the rule-building agent degrades if
%  prompted with large samples of data}. Since bespoke
% discrete search can use all the data available, this handicaps LLM-based %systems.
%  We show that this disadvantage can be compensated for by (1) giving CLIPPER
%  access to tools for computing certain statistics over the full
%  dataset or even (2) incorporating some of these statistics into the
% prompt given to the coding agent.  Even without these extensions, however,
%{CLIPPER outperforms SLIPPER on smaller training sets} (up to a few hundred %examples---approximately 5x the sample size we used).

\section{Related work}
\label{sec:related}

RIPPER combines an outer rule-set construction procedure with discrete rule
search and pruning \citep{cohen1995ripper}. SLIPPER replaces the conventional
set-covering outer loop with confidence-rated boosting, while retaining a
greedy rule builder related to RIPPER's inner loop
\citep{cohen1999slipper,schapire1999confidence}. A rule contributes a fixed
confidence to the examples it covers and abstains on other examples; the
ensemble combines the contributions of the rules that fire.

\todo{Expand the comparison with interpretable classifiers, rule ensembles,
LLM-based classifier ensembles (including the reference to Zico's paper in
the outline), and agent harnesses. Add the corresponding citations and
clarify how the final hypothesis differs from these approaches.}

% Close the group left open in the supplied source, outside the source file.
}
\endgroup
\section{Related work}
\label{sec:related}
\textbf{Symbolic classifiers.}  We build on, and compare to, prior symbolic learners, in particular rule learners.  Our work is most closely related to 
RIPPER, which uses an effective but efficient search for a compact rule set
\citep{cohen1995ripper}, and
SLIPPER, which uses confidence-rated boosting
\citep{cohen1999slipper,schapire1999confidence} to build a comparably small ensemble of weighted rules.  We use our own re-implementation of SLIPPER\footnote{A Python rewrite of the original C implementation.} and CLIPPER shares most of the code, but replaces the discrete rule search with a call to a Pydantic AI\footnote{\url{https://pydantic.dev/}} agent.  We compare to other models, including CORELS, which searches for an optimal
regularized rule list over a fixed set of antecedents
\citep{angelino2018corels}; RuleFit, which is a hybrid of rule models and linear classifiers \citep{friedman2008rulefit};  FIGS, which learns an ensemble of small
trees \citep{tan2025figs}; and others. Many of these comparisons were facilitated by the
\texttt{imodels} package \citep{singh2021imodels}.

\textbf{LLM-generated features for downstream  symbolic learners.}
Past work \citep{singh2023augimodels,han2024featllm,abhyankar2025llmfe,
zhang2025llmlasso,jia2026tabdistill} uses LLMs to augment or select features for symbolic learners.  One well-known example is FeatLLM \citep{han2024featllm}, which uses bagging and prompting in a few-shot setting to generate many
conjunctive rules (much like the ones produced by SLIPPER), which are used by a symbolic learner.  CLIPPER differs in generating a compact classifier directly, similarly to SLIPPER: e.g., in most of the experiments from \citet{han2024featllm}, 200 rules are generated for each class, from up to 64 examples, while CLIPPER generates an average of 9.6 rules on 2000 examples, and SLIPPER generates an average of 8.9.   

\textbf{LLMs for interpretable model construction.}
A zero-shot decision tree learner is described by \cite{carrasco2025zeroshotdecisiontreeconstruction}, where the learner is modified so that 
data-derived statistics are replaced with LLM-driven decisions.  Our work is similar in spirit but combines LLM background knowledge with data.  In concurrent work, RuleChef describes an agentic system that generates executable rules
for text classification and extraction \mbox{\citep{kovacs2026rulechef}}. We compare to it
in Section~\ref{sec:rulechef} and Appendix~\ref{app:banking77}. Unlike RuleChef, CLIPPER is evaluated on tabular data, and our framework allows controlled evaluation of an LLM's ability to produce rules from data.

\textbf{Boosting language-model classifiers.}
\citet{manikandan2023weak} use LLMs as weak learners on tabular
data: each round produces a natural-language summary, which conditions an
LLM when predicting labels. CLIPPER also uses boosting to combine weak learners, but each weak
hypothesis is executable code, so prediction needs no LLM.

% seems like a stretch but if we have space...
%\textbf{Code generation with feedback.}
%SWE-agent shows that the interface for inspecting, editing, and executing
%code affects a coding agent's performance \citep{yang2024sweagent}.
%FunSearch combines LLM program proposals with an automated evaluator
%\citep{romeraparedes2024funsearch}. CLIPPER also separates proposal from
%evaluation, using weighted training coverage to evaluate candidate rules,
%but its target is generalization to held-out examples.

\section{Method}
\label{sec:method}

\begin{figure*}[t]
\centering
\includegraphics[width=\textwidth]{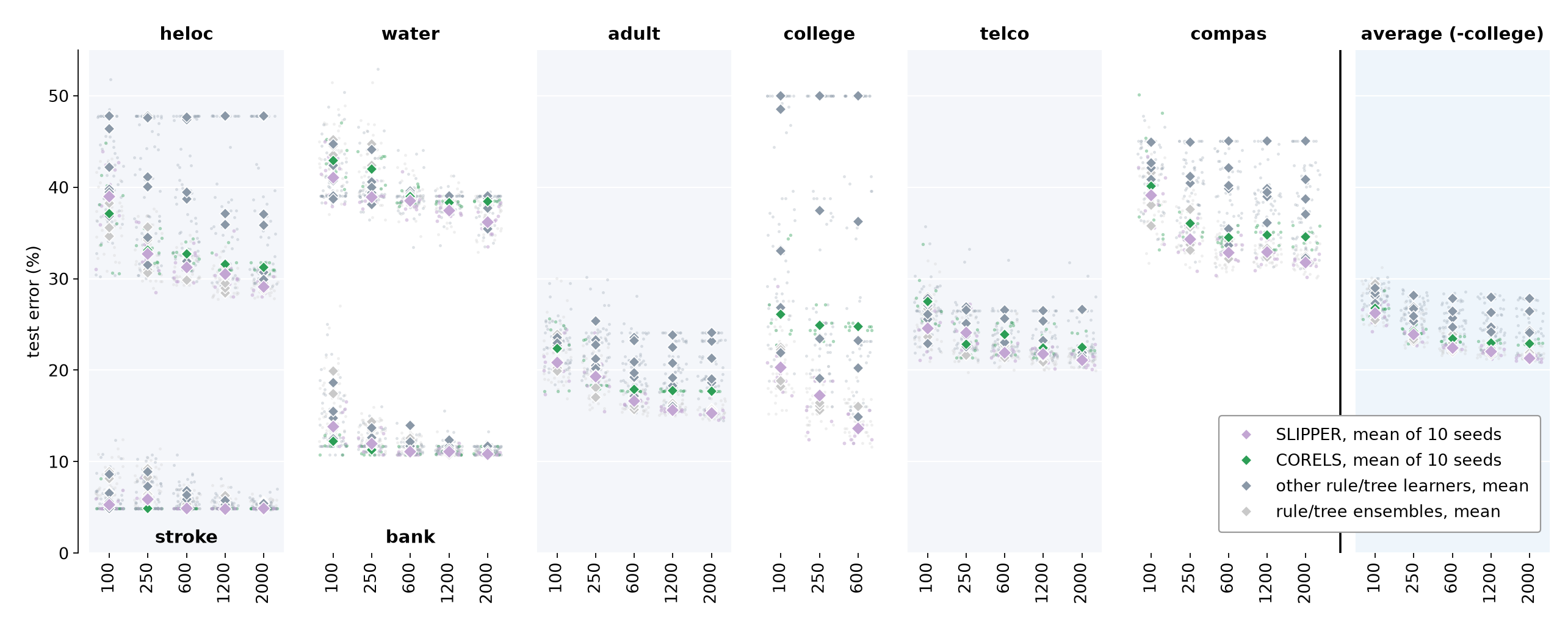}
\caption{\textbf{SLIPPER is a strong symbolic baseline.} On 4,510 experiments, SLIPPER (purple) and CORELS (green) are the two strongest rule learners. Weaker rule and tree
  learners are blue; ensembles are gray.  The
  shaded band (right) is average performance over the 7 larger benchmarks. In the first 6 bands the faint dots are 
  individual runs with different seeds (jittered to reduce
  overlap) and the diamonds are means; in the shaded band each dot is
  one seed's average over the 7 benchmarks.  Benchmarks with
  distinct error ranges are stacked to save space.}
\label{fig:all-baselines}
\end{figure*}

Before deciding to develop CLIPPER we evaluated 12 interpretable learners on 8
tabular benchmarks previously selected by other authors \citep{kruschel2026challenging} as being appropriate tasks for use of interpretable models (e.g., COMPAS estimates the likelihood of recidivism).
We constructed binary versions of each of these datasets, and evaluated error rates at up to five training-set sizes.\footnote{\emph{college}, with only 750 training instances, uses only 3 sizes.}  Running 10 seeds for each method gives 4510 experiments, shown in Figure~\ref{fig:all-baselines}, and SLIPPER emerged as the strongest baseline learner, followed by CORELS.

SLIPPER's learning method lends itself to
adaptation to a harness, which we describe as follows, letting $(x_i,y_i)_{i=1}^{n}$ be a binary training set with $y_i\in\{-1,+1\}$,
and letting $D_t(i)$ be the normalized weight of example $i$ at round $t$.  For $t=1$, $D_t$ is uniform.

\emph{Finding a rule.} 
In round $t$ SLIPPER uses a greedy discrete search to find a ``rule'' $r_t$, i.e., a learned predicate $r_t(x)\in\{0,1\}$ that ``covers'' an example when it returns one. 
The rule's positive coverage $p_t$, negative coverage $q_t$, and weight $\alpha_t$ are computed as
\begin{align}
 p_t &= \sum_{i:y_i=+1}D_t(i)r_t(x_i), &
 q_t &= \sum_{i:y_i=-1}D_t(i)r_t(x_i), & 
  \epsilon &=\frac{1}{2n}, &
 \alpha_t &= \tfrac12\log\frac{p_t+\epsilon}{q_t+\epsilon}.
 %G(p_t,q_t)&=\sqrt{p_t}-\sqrt{q_t},  \qquad
 \label{eq:rule-score}
\end{align}

\emph{Reweighting the examples.}  The rule weight $\alpha_t$ is used to update the example weights:
\begin{equation}
 D_{t+1}(i) \propto  D_t(i)\exp[-y_i \alpha_t r_t(x_i)] %{Z_t}%,
 %\qquad F(x)=b+\sum_{t\in\mathcal{A}}\alpha_t r_t(x),
 \label{eq:boost}
\end{equation}

\emph{Rejecting rules and the default rule.}  On round 1, SLIPPER defines the ``default rule'' $r_1(x) \equiv -1$ (for any $x$).  The other rules always return 0 or 1, so this is the only rule that recommends the negative class.
In subsequent rounds, SLIPPER's search tries (roughly speaking) to maximize $(\sqrt{p_t}-\sqrt{q_t})$ without overfitting.  If this quantity is negative then $r_t$ is replaced with $r_1$ and its weight $\alpha_1$ is incremented by the appropriate $\alpha_t$.  The final classifier combines the rule weights and predicts the sign of $F(x)$:
$$ F(x)=\sum_{t}\alpha_t r_t(x)=b+\sum_{t\in\mathcal{A}}\alpha_t r_t(x)
$$
 where $b$ is a bias term (the contribution of $r_1$) and  $\mathcal{A}$ indexes accepted rules.  

\emph{SLIPPER's rules are sparse.}  Importantly \textbf{$F(x)$ only depends on the rules that cover $x$}, unlike the case for ordinary boosting, where all weak hypotheses contribute to every prediction.

\emph{Boosting as a harness and hyperparameters.}  \textbf{CLIPPER uses this same procedure, except that it invokes an LLM agent to build a rule.}  One hyperparameter of SLIPPER (and thus CLIPPER)  is the number of rounds of boosting $T$.\footnote{$T$ upper-bounds the number of rules learned---the number of rules will be less than $T$ if rules are rejected.}  SLIPPER bounds $T<100$, and uses 5-fold internal cross-validation to pick the best value $T^*$.  CLIPPER is similar, but to reduce LLM usage it cross-validates SLIPPER (not itself) to pick $T^*$: the only exception is that because
CLIPPER's rule builder is more productive, it always allows at least two rounds of boosting, even when SLIPPER's cross-validation prefers 0 or 1.
  \textbf{All other hyperparameters for CLIPPER were set on an independent tuning set of benchmarks.}  That includes the prompts, the tools, and other configuration choices.\footnote{We developed CLIPPER on a disjoint suite of ten binary tasks: Abalone,
Breast cancer Wisconsin, Cylinder bands, German credit, Hypothyroid, Move1,
Obesity, Pima diabetes, Portuguese student performance, and Titanic
(Appendix~\ref{app:development}).} Hence in the results of Section~\ref{sec:experiments}, \textbf{every reported result measures performance
on tasks that played no part in designing the method.}

Appendix~\ref{app:protocol} gives the full procedure.

\emph{Details of the prompting and the tools.}  The agent receives a weighted sample of at most $M$ examples, drawn without
replacement from $D_t$, together with labels, sampled weights, and the tabular data schema.
For tabular data, we use $M=100$ and a Markdown table with quoted string
values to encode the data sample.  Note that a different sample is passed to the LLM in each boosting round, so
potentially many more than $M$ can be seen by CLIPPER during learning.  We also modified SLIPPER to
use samples of size $M$ in order to perform controlled experiments (below).
The prompt asks for one conjunctive rule implemented as a Python function
\texttt{covers(x)}, where $x$ is an attribute dictionary. This restriction comes from prompting: the executor accepts more general Python predicates.

\emph{Variants.} We compare four main CLIPPER variants. \emph{Sample} provides the examples and
schema alone. \emph{Coverage} adds a tool that, given a predicate, returns values of $p_t$ and $q_t$ computed on the full
training set. \emph{Hints} adds to the prompt a table of the best single condition
for each attribute, as found by SLIPPER's discrete optimizer, with sample and
full-data coverage. \emph{Hints+tools} also adds this hint output to the output of the coverage tool. Coverage uses a six-call
tool limit per attempt; hints+tools uses two. All conditions receive the same
full-data scoring by the harness after a rule is found. %The sample-only comparison
%therefore matches direct example access, while hints and tools expose
%additional training statistics.

%\subsection{Text classification and value extraction}

\emph{Textual data and multiple classes.}  On tabular data, SLIPPER is restricted to rules containing conjunctions of equality tests for discrete attributes and inequality tests for continuous ones; it also can construct tests on the presence of sparse attributes (e.g., words, for classifying text).  Using LLM rule builders allows generation of many other types of rules by slightly varying the prompt.  We consider string operations and regular expressions for classification.  We also consider rules that extract data from text. For extraction, a rule returns a value or abstains. In
Equation~\ref{eq:rule-score}, $p_t$ becomes the weight of correct returned
values and $q_t$ the weight of incorrect returns; abstentions contribute to
neither. We follow the same weight update procedure as in classification; the default is abstention, and a rule is retained only
if $p_t>q_t$. At prediction time, confidences are summed for each canonical
value and the largest sum wins. If no rule matches, the system abstains.
We compare unrestricted Python extraction functions with structured regex
rules containing patterns, flags, and a capture-group template.

%Weighted rule selection also applies to richer rule languages; for text
%classification, predicates operate on a raw patient note (mirroring realistic clinical calculation tasks) and may use string operations
%or regular expressions. We combine binary ensembles by
For multiclass problems we use one-versus-rest prediction (Section~\ref{sec:medcalc}).

\section{Experiments}
\label{sec:experiments}

% OLD: We compare the coding agent with discrete rule search at equal information,
% OLD: extend it to more data and to established baselines, separate the roles of
% OLD: harness and model, and extend the rule language to text.
We compare the coding agent with discrete rule search at equal information,
extend it to more data, compare it with established and LLM-based rule
learners on tabular data and text, and separate the roles of harness and model.

\subsection{Evaluation protocol}

The tabular evaluation uses eight binary tasks drawn from the suite
studied by \citet{kruschel2026challenging}: Adult, Bank marketing, College,
COMPAS, HELOC, Stroke, Telco churn, and Water potability. Each has a fixed,
stratified 25\% test holdout. For each training size and seed, methods receive
the same training subset. As noted above, CLIPPER and SLIPPER use the same boosting budget (except CLIPPER always boosts at least twice). The default coding model is Gemini~3.1 Flash-Lite, a fast, inexpensive model, with default generation hyperparameters.

Results on tabular datasets report the mean test-error percentage and standard error of the mean (SEM) across 10 seeds (unless noted otherwise). Macro error averages dataset means equally, with propagated SEM. Learning curves that extend past $n=750$ exclude college, which has only 750 training examples.
Appendix~\ref{app:protocol} specifies additional details.

\subsection{LLMs build better rules than discrete search}
\label{sec:tabular}

\begin{figure}[t]
\centering
\includegraphics[width=\linewidth]{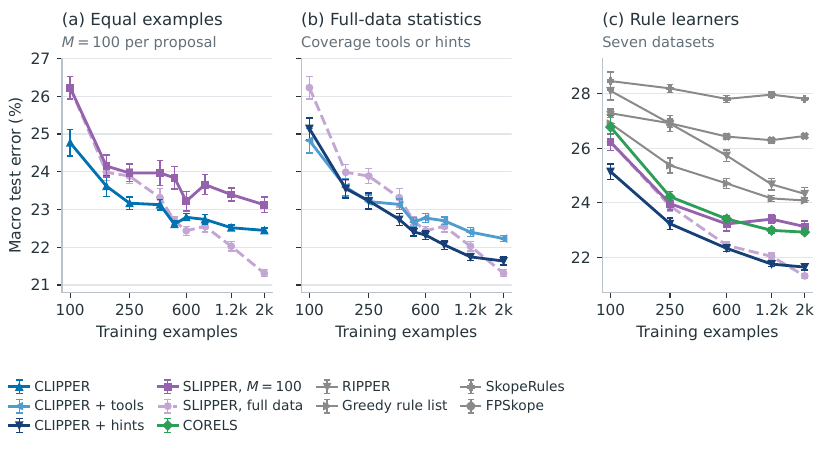}
% OLD: \caption{\textbf{Why a coding agent helps.} (a,b) Macro test error on
% OLD: seven tabular datasets (mean and propagated SEM, ten seeds). CLIPPER sees at
% OLD: most $M=100$ examples per proposal, and every proposal is scored on the full
% OLD: training set. (a) CLIPPER against SLIPPER given the same examples ($M=100$)
% OLD: and given all training rows. (b) Full-data condition hints let CLIPPER use
% OLD: more of the training set. (c) Paired error changes when dataset, feature,
% OLD: class, and value names are replaced with arbitrary identifiers (CLIPPER with
% OLD: coverage tools, $n=250$, all eight datasets). Grey dots are seeds; Average
% OLD: shows each seed's mean change over the eight datasets. Boxes show quartiles and medians with
% OLD: $1.5$-IQR whiskers. Stars mark paired $t$-test significance (*~$p<0.05$,
% OLD: **~$p<0.01$, ***~$p<0.001$); green boxes are significant increases, gray
% OLD: boxes not significant.}
\caption{\textbf{CLIPPER on tabular data.} Macro test error on
seven tabular datasets (mean and propagated SEM, ten seeds). CLIPPER sees at
most $M=100$ examples per proposal, and every proposal is scored on the full
training set. (a) CLIPPER against SLIPPER given the same examples ($M=100$)
and given all training rows. (b) CLIPPER given full-data statistics in two
ways: coverage tools or condition hints. (c) CLIPPER with hints against
established rule learners.}
\label{fig:tabular-results}
\end{figure}

With both rule builders limited to a sample of $M=100$ examples drawn in each boosting round, CLIPPER
has lower macro test error than SLIPPER at all nine training sizes
(Figure~\ref{fig:tabular-results}a); at $n=1{,}200$, CLIPPER's error is
$0.9\pm0.2$ points lower (22.5 versus 23.4\%). In this controlled experiment, the only difference is the way that rules are built: by statistics and discrete search for SLIPPER, and by LLMs for CLIPPER.

%\subsection{Using more data without longer prompts}
\label{sec:hints}

Of course, SLIPPER's rule-building does not need to limit itself to a sample of 100, as the dashed line in Figure~\ref{fig:tabular-results}a shows.  One might hope that giving the LLM a larger sample would help, but this is \emph{not} true.\footnote{On the development suite,
increasing the prompt from $M=100$ to 200 or 400 examples \emph{raised} macro error slightly (by 0.3--0.4 points; Appendix~\ref{app:development}). On
MedCalc (Section~\ref{sec:medcalc}), which has very large examples, CLIPPER with $M=20$ is more accurate
than with $M=100$ (Table~\ref{tab:medcalc-final}). Development suite experiments suggest that the Gemini model we use as a base gets less accurate at estimating dataset statistics as the sample grows.}  However, giving direct access to full-data statistics improves performance substantially, especially with more training data.  The ``hints'' variant of CLIPPER keeps the same sample size, but adds to the prompt the best single condition for each attribute, as found by SLIPPER's discrete search. Without hints, CLIPPER improves over full-data SLIPPER up to about $n=500$, about 5 times the sample size. 
With hints, CLIPPER beats full-data SLIPPER at eight of nine sizes
(Figure~\ref{fig:tabular-results}b). %At $n=1{,}200$, hints reach
%21.7\%, compared with 22.5\% for sample-only CLIPPER
%and 22.0\% for full-data SLIPPER. 
Coverage
queries alone give smaller gains (Figures~\ref{fig:tabular-results}b and~\ref{fig:tabular-all}).

%\subsection{Comparison with established rule learners}
\label{sec:rule-learners}

% OLD: Figure~\ref{fig:rule-learners-calculator}a compares CLIPPER with coverage
Figure~\ref{fig:tabular-results}c compares CLIPPER with hints
against the other rule learners from Figure~\ref{fig:all-baselines}.
CLIPPER has lower error than CORELS, RIPPER, the greedy rule list,
SkopeRules, FPSkope, and sampled SLIPPER at all five training sizes; only
full-data SLIPPER is lower, and only at $n=2{,}000$. CLIPPER's learned rulesets are also small: across CLIPPER conditions
and training sizes, the median ruleset contains 4.5 to 13.5 rules, well within
the range humans can inspect (\citealp{Heider2023explainableAI};
Appendix~\ref{app:model-size}).

\begin{figure}[t]
\centering
\includegraphics[width=\linewidth]{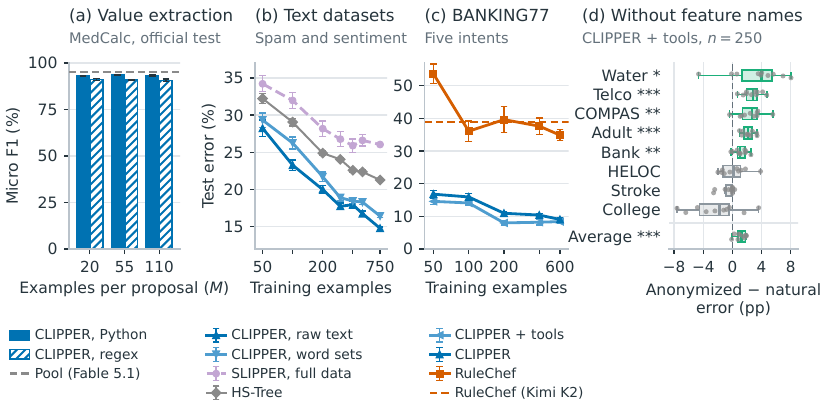}
\caption{\textbf{CLIPPER on text.} Bars and points show means over ten
seeds, with SEM error bars, propagated across datasets in (b); CLIPPER uses Gemini~3.1 Flash-Lite throughout. (a) Official-test value extraction
(Section~\ref{sec:extraction}): micro F1 of Python and regex rules learned
without coverage tools. Dashed: the reference pool, written with
Claude Fable~5.1 and evaluated once. % OLD: (b) Exact 55-way calculator prediction
% OLD: from patient notes on MedCalc-Bench-Verified's 1,100-row official test split.
% OLD: CLIPPER reads the notes and SLIPPER their word sets; neither receives the
% OLD: question. Sampled learners use at most $M=20$ examples per proposal;
% OLD: full-data SLIPPER uses all training rows.
(b) Macro error on five text datasets. CLIPPER uses no coverage tools and
reads either the raw text or, like SLIPPER, word sets. (c) Test error on five BANKING77
intents (200 official test queries). CLIPPER and RuleChef read the raw text
with the same model and training queries. Dashed: RuleChef's reported error
with Kimi K2 \citep{kovacs2026rulechef}, counting unlabeled queries as errors
(our runs give them the most common intent). (d) Paired error changes when dataset, feature,
class, and value names are replaced with arbitrary identifiers (all eight tabular datasets). Grey dots are seeds; Average
is each seed's mean over the eight datasets. Boxes: quartiles, medians,
$1.5$-IQR whiskers. Stars mark paired $t$-test significance (*~$p<0.05$,
**~$p<0.01$, ***~$p<0.001$); green boxes are significant increases.}
\label{fig:text-results}
\end{figure}

\textbf{Text classification.}\label{sec:medcalc}
% OLD: We adapt MedCalc-Bench-Verified \citep{khandekar2024medcalc} to a 55-class
% OLD: task: predicting which calculator an example requires from its patient note
% OLD: alone, without the question, which names the calculator. SLIPPER cannot read
% OLD: free text, so it sees each note as the set of lowercase words it contains,
% OLD: without numbers or word order (Appendix~\ref{app:medcalc}). CLIPPER instead
% OLD: writes Python predicates over the raw note, with string operations, regular
% OLD: expressions, and coverage tools. We fit 55 one-versus-rest ensembles for
% OLD: six rounds each and predict from their centered scores. With $M=20$ examples
% OLD: per proposal, CLIPPER's error is 5.3 to 6.4 points lower than sampled
% OLD: SLIPPER's from 220 training notes on, and below full-data SLIPPER's at 220
% OLD: and 880 (Figure~\ref{fig:text-results}b).
We also constructed a benchmark suite with five binary text-classification
tasks, two for spam detection (SMS spam, YouTube spam) and three for sentiment
(Amazon, IMDb, and Yelp reviews; Appendix~\ref{app:text-suite}). We use CLIPPER without tools or
hints, with two prompt variants: one that directs the LLM to construct the
same sort of rule conditions as are supported by SLIPPER (``word sets'', which
allow testing if a token appears) and one that allows regular expressions on
the raw text. We compare to SLIPPER with the same conditions and no sampling.
Here both variants of CLIPPER substantially improve on SLIPPER, and also on
the best baseline learner (in this case HS-Tree, an \texttt{imodels} tree
learner with hierarchical shrinkage, using TF-IDF-transformed features;
Figure~\ref{fig:text-results}b). CLIPPER also does well on a 55-way
task: predicting which MedCalc-Bench-Verified calculator
\citep{khandekar2024medcalc} a patient note requires
(Appendix~\ref{app:medcalc}).

\label{sec:rulechef}Recent systems also use an LLM to write rules. The closest to CLIPPER,
RuleChef \citep{kovacs2026rulechef}, asks an LLM to write rules from labeled
examples, but without boosting: it writes a small set of rules and then
revises them to fix their mistakes. We compare the two on the benchmark
RuleChef's authors use: five intents from BANKING77
\citep{casanueva2020efficient}, with all 200 official test queries for those
intents (Appendix~\ref{app:banking77}). Both methods read the raw text and use
the same model, Gemini~3.1 Flash-Lite, and RuleChef runs its authors' own
benchmark script and settings. CLIPPER fits one-versus-rest models as for
MedCalc, with $M=100$. CLIPPER's error is lower than RuleChef's in all
50 runs (five training sizes, ten seeds each), by 29.7 points on average with
coverage tools and 27.8 points without them
(Figure~\ref{fig:text-results}c). From $n=100$ on, RuleChef's error stays
between 34.9 and 39.7\%, while CLIPPER with tools reaches 8.0\% at $n=200$.
RuleChef's rules label only about half of the test queries: after its first
rounds of revision, it usually replaces its initial rules with one broad
regular expression per intent.

% \todo{1 - why use the tool-clipper in 3c and the hint-clipper in 3b?
% 2 - I would move 4a to the appendix (sorry, I know you worked hard on it) and squeeze in Table 4, or some variant of it - that's a much stronger result and we should lead with it more heavily. 3 - should drop rulechef from 4b.

% }

\textbf{Background knowledge.}\label{sec:new-anonymization}\label{sec:knowledge}
We believe that CLIPPER's ability to leverage its knowledge of language is why it has such strong performance on these text datasets. Reading the same word sets as full-data SLIPPER,
CLIPPER without tools has lower error at every size (29.3 versus
34.2\% at $n=50$; 16.4 versus 26.0\% at $n=750$;
Figure~\ref{fig:text-results}b). On a text dataset whose labels were published after
the model's training cutoff, CLIPPER reading the raw text still has the lowest
error at $n=750$
(6.9\%, versus 8.4\% for SLIPPER; Appendix~\ref{app:text-suite}). CLIPPER's edge over earlier rule learners on tabular data is also likely connected to its use of LLM background knowledge.
To test this, we replace all the dataset, feature, class, and
categorical-value names with meaningless identifiers in the tabular datasets. At $n=250$, macro error on all eight datasets rises from
22.7 to 23.8\%, a paired increase of $1.1\pm0.2$
% OLD: points (Figure~\ref{fig:proposal-information}c). Semantic cues help, but
points (Figure~\ref{fig:text-results}d). %Semantic cues help, but
%CLIPPER retains surprising competence without them: on the seven datasets of
% OLD: Figure~\ref{fig:rule-learners-calculator}a, anonymized CLIPPER
%Figure~\ref{fig:tabular-results}c, anonymized CLIPPER
%(24.8\%) still outperforms RIPPER, the greedy rule list,
%SkopeRules, and FPSkope using all training rows.

\subsection{From predicates to value-returning rules}
\label{sec:extraction}

A text rule can also return a value. We supply the calculator and requested
entity key and ask the learner to extract its value from the note, for the
ten most frequent entity keys (Appendix~\ref{app:medcalc}). Each key has its own ten-round ensemble. Python and regex rules use
$M=20,55,110$, with ten seeds and no coverage tools. As a reference, we
built a pool of regex rules by inspecting the training data with Anthropic's
Claude Fable~5.1 as our coding agent, then selected rules from this fixed
pool with the same scorer. We evaluate on 500 official-test calculator-row
instances containing 1,120 values.

With a much weaker model, CLIPPER learns extractors that approach those
written with a frontier coding agent. Python rules learned with Gemini~3.1
Flash-Lite at $M=55$ reach 93.7 micro F1, within 1.6 points of the
reference pool's 95.3 (Figure~\ref{fig:text-results}a), although
Flash-Lite trails Fable~5.1 substantially on published benchmarks
\citep{google2026flashlitecard,anthropic2026fable51}. The richer rule
language helps: Python and regex rules are similarly precise (96.1
and 96.3\%), but Python recovers more of the requested values
(recall 91.4 versus 86.0\%;
Table~\ref{tab:extraction-final}).

\subsection{The harness matters more than the model}
\label{sec:harness}

\textbf{One rule at a time.}
We first replace iterative rule construction with a one-shot prompt for a
complete positive-class classifier, keeping the LLM model, sample cap, and tools
fixed (Table~\ref{tab:harness-main}). One-shot error is higher than
boosted CLIPPER's at $n=250$ (23.9 versus 23.2\%), and
the gap widens with more data: at $n=1{,}200$, one-shot error rises to
24.8\%, while boosting falls to 22.4\%. 

Another common way of learning rules iteratively is by ``set covering'': after each rule is learned, all examples it covers are removed, and a new rule is learned from the remaining data, approximating a greedy set-covering algorithm.  Surprisingly, while replacing boosting with set covering negatively affects SLIPPER, the same harness change is sometimes slightly better when using the LLM-based rule builder.

%which then has lower error than full-data discrete set cover
%at every size. Iterative construction, rather than boosting specifically,
%is essential.

\begin{table}[t]
\caption{\textbf{Iterative construction matters; boosting specifically does
not.} Macro test error (\%) on seven tabular datasets, mean $\pm$ propagated
SEM over ten seeds. The LLM rule builder uses coverage tools and sees at most
$M=100$ examples per proposal; discrete search uses all training rows.
Discrete set cover is a repaired variant selected on the development suite
(Appendix~\ref{app:setcover}).}
\label{tab:harness-main}
\centering
\small
\setlength{\tabcolsep}{4pt}
% Hand-made from source/results/tabular.csv, repaired-cover.csv, and
% repaired-cover-more-sizes.csv; set cover uses patience 1 (CLIPPER) and the
% development-selected repaired variant (discrete search).
\begin{tabular}{@{}llccccc@{}}
\toprule
Harness & Model / rule search & $n=100$ & $n=250$ & $n=600$ & $n=1{,}200$ & $n=2{,}000$ \\
\midrule
Boosting & Gemini 3.1 Flash-Lite & $24.8\pm0.3$ & $23.2\pm0.2$ & $22.8\pm0.1$ & $\boldsymbol{22.4\pm0.1}$ & $\boldsymbol{22.2\pm0.1}$ \\
Set cover & Gemini 3.1 Flash-Lite & $\boldsymbol{24.4\pm0.3}$ & $\boldsymbol{23.0\pm0.1}$ & $\boldsymbol{22.5\pm0.1}$ & $22.5\pm0.1$ & $22.4\pm0.1$ \\
One-shot & Gemini 3.1 Flash-Lite & $25.4\pm0.4$ & $23.9\pm0.3$ & $24.1\pm0.3$ & $24.8\pm0.4$ & $24.2\pm0.4$ \\
\midrule
Boosting & discrete search & $\boldsymbol{26.2\pm0.3}$ & $\boldsymbol{23.9\pm0.2}$ & $\boldsymbol{22.4\pm0.1}$ & $\boldsymbol{22.0\pm0.1}$ & $\boldsymbol{21.3\pm0.1}$ \\
Set cover & discrete search & $26.7\pm0.4$ & $24.6\pm0.2$ & $23.4\pm0.1$ & $22.8\pm0.1$ & $22.6\pm0.1$ \\
\bottomrule
\end{tabular}

\end{table}

\begin{table}[t]
\footnotesize
\setlength{\tabcolsep}{3pt}
\begin{minipage}[t]{0.61\linewidth}
\centering
\caption{\textbf{General-purpose coding agents.} Macro test error (\%)
on seven tabular datasets at $n=250$, mean $\pm$ propagated SEM over three matched
seeds. Runs in which an agent fit an existing learner were detected and
corrected (Appendix~\ref{app:new-agent-fitting}).}
\label{tab:agents}
% Hand-made from tables/new-harness-full.tex (Macro 7 column).
\begin{tabular}{@{}lllc@{}}
\toprule
Harness & Model / rule search & Tools & Error (\%) \\
\midrule
Codex & \textbf{Sol} & default & $\boldsymbol{22.9\pm0.2}$ \\
 & Terra & default & $23.5\pm0.3$ \\
 & Luna & default & $24.2\pm0.4$ \\
\addlinespace[2pt]
Claude Code & Opus 5 & default & $23.0\pm0.4$ \\
 & Sonnet 5 & default & $23.6\pm0.3$ \\
 & Haiku 4.5 & default & $28.1\pm1.1$ \\
\addlinespace[2pt]
Gemini CLI & Gemini 3.5 Flash & default & $24.0\pm0.3$ \\
 & Gemini 3.1 Flash-Lite & default & $26.8\pm0.9$ \\
\midrule
Boosting & \textbf{Gemini 3.1 Flash-Lite} & coverage & $\boldsymbol{23.5\pm0.4}$ \\
Boosting & discrete search & --- & $24.3\pm0.4$ \\
\bottomrule
\end{tabular}

\end{minipage}\hfill
\begin{minipage}[t]{0.36\linewidth}
\centering
\caption{\textbf{Changing CLIPPER's model.} Boosting without coverage
tools, $n=250$, three matched seeds (Appendix~\ref{app:new-backend-details}).}
\label{tab:backends}
% Hand-made from tables/new-backend-full.tex (fixed-seven column, without coverage tools).
\begin{tabular}{@{}lc@{}}
\toprule
Model & Error (\%) \\
\midrule
\textbf{Sonnet 5} & $\boldsymbol{23.3\pm0.4}$ \\
Gemini 3.1 Flash-Lite & $23.4\pm0.4$ \\
Gemini 2.5 Flash-Lite & $23.9\pm0.3$ \\
Haiku 4.5 & $24.2\pm0.4$ \\
Qwen 3.5 9B & $24.8\pm0.5$ \\
Gemini 3.5 Flash & $24.8\pm0.6$ \\
\bottomrule
\end{tabular}

\end{minipage}
\end{table}

\textbf{Changing the model.}\label{sec:new-backends}
Replacing the model inside CLIPPER changes error little
(Table~\ref{tab:backends}): all six models, including the small open-weight
Qwen~3.5~9B, fall within 1.5 points. The model used for development, Gemini~3.1 Flash-Lite, nearly matches the best
model, Sonnet~5, at a small fraction of its cost. 

\textbf{General-purpose coding agents.}\label{sec:new-general-agents}

Changing the harness and fixing the model leads to larger changes in performance.  We also gave the whole learning task to general-purpose coding agents,
which can use their built-in tools to inspect all 250 training examples, write and execute code, and so on (Table~\ref{tab:agents}; Appendix~\ref{app:new-agent-protocol}).
Frontier models do well with their corresponding coding harnesses, but small models do not:
Gemini~3.1 Flash-Lite and Haiku~4.5 are 3--4 points worse in their coding harnesses than
inside CLIPPER (Tables~\ref{tab:agents} and~\ref{tab:backends}). Inside CLIPPER,
Flash-Lite falls within the range of the frontier agents and beats
full-data SLIPPER, at about \$0.04 per learning run, versus
\$0.39--1.02 for Terra, Sonnet, Gemini~3.5, and Opus.

Qualitatively, the general-purpose coding agents behave quite differently than CLIPPER, as might be expected.  They do not simply write rules in code: instead, they often write code for discrete search methods and use the results.  In several cases agents tried to download and run existing interpretable learners (Appendix \ref{app:new-agent-behavior}).\footnote{We attempted to discourage this with our prompts, and believe that the results above are not from use of other baselines.  While ways of combining the output of existing interpretable learners may be of interest, it is not the main focus of this work.}  We also note that some additional normally-strong coding models were tried but discarded because of ``overthinking'': i.e., taking many minutes and emitting thousands of thinking tokens on a single boosting round.

\section{Discussion and conclusion}
\label{sec:discussion}

% \todo{this is pretty different so take a look}

Our main interest in this work is understanding what LLMs can and cannot do.
Although writing predictive rules is different from existing programming benchmarks, LLMs can be used to produce predictive rules, and combined with simple harnesses based on known rule-learning methods, can directly construct accurate interpretable classifiers.  This result extends past work in LLM-driven construction of features to the harder problem of constructing complete classifiers, and using LLMs to write mechanistically interpretable executable classifiers provides a way of distilling LLM knowledge into these classifiers in a data-guided way.

An important outcome of this work is a new and better rule learner.
This is of interest because interpretable models are a useful alternative to black-box models in some practically important settings---among them, auditing the behavior of LLMs! 

Given the same data and same constraints, 
CLIPPER writes better rules than discrete search, on tabular data, where it is consistently better than or competitive with even full-data SLIPPER, and on
% OLD: clinical notes. It draws
text, where it wins by large margins. %It draws on its priors about what the features mean, yet preserves surprising capability in their absence through the ingenuity of the rule learning mechanism
%(Figure~\ref{fig:text-results}d). The classic harness thus complements and extends the model inside it: with 
With the CLIPPER harness, an inexpensive model performs comparably to general-purpose agents built on frontier models, at a
fraction of their cost. %, while the same model loses 3 to 4 points when left %to
%build a classifier on its own (Table~\ref{tab:agents}). 
Using an LLM extends classical rule-learning algorithms beyond fixed attribute tests, and by doing so, we are able to construct text extractors with a small model that are highly competitive with extractors written interactively by a developer in collaboration with a frontier model.

% OLD: CLIPPER's main limitation also has a classical remedy. A prompt holds
% OLD: only a sample, and larger samples do not help, so full-data SLIPPER
% OLD: eventually overtakes sample-only CLIPPER. Giving the LLM the best single
% OLD: conditions found by SLIPPER's discrete search closes this gap: with these
% OLD: hints, CLIPPER beats full-data SLIPPER at eight of nine training sizes.
% OLD: Coding LLMs do not replace symbolic learners; they are most effective when
% OLD: built on them. Finally, learning interpretable classifiers is a demanding
% OLD: test for coding agents: held-out error measures quality directly, mature
% OLD: learners set a strong baseline, and the harness, more than the model,
% OLD: decides whether a small model succeeds.
Two natural next steps are fine-tuning small models to write better
rules, with the score the harness already gives every proposed rule as a
training signal, and scaling CLIPPER to larger samples of the training data.
Both could make small, inexpensive models even stronger rule learners. 
Finally, learning interpretable classifiers is a useful benchmark for coding
agents: it measures quality by held-out error, compares against mature
specialist algorithms, and produces interpretable programs.

\iffalse

The same lesson answers the agent's one limitation. A prompt holds only a
sample, and larger samples do not help, so full-data SLIPPER eventually
overtakes sample-only CLIPPER. The remedy again comes from classic rule
learning: discrete search computes the best conditions over every training
row, and the agent composes them into rules. With these hints, CLIPPER beats
full-data SLIPPER at eight of nine training sizes
(Figure~\ref{fig:tabular-results}b). Coding agents do not replace
symbolic learners; they are most effective when built on them.

Finally, learning interpretable classifiers is a useful benchmark for coding
agents: it measures quality by held-out error, compares against mature
specialist algorithms, and produces programs that can be inspected directly.
Because the agent writes code rather than searching a fixed space of
attribute tests, the same harness can learn model classes that no discrete
rule learner supports. We see this as the most promising direction: pairing
the statistical discipline of classical rule learning with what coding agents
know and can express, and allowing for more efficiently designed, interpretable classifiers for any task.
\fi

% Staged proposals for review, after the conclusion.
% \clearpage
% \input{sections/new}

% Flush main-text figures before the end matter. Keep ICLR's layout unchanged.
% The arXiv preprint has no page limit; its statements follow the conclusion.
\ifdefined\ARXIV\else\clearpage\fi
\subsection*{AI use statement}

We use an LLM to generate executable rules during CLIPPER training, as
described in Section~\ref{sec:method}. We also used AI assistants to help
find related work, plan and interpret experiments, construct the hand-written
extraction pool, write and inspect parts of experiment code and outputs, and
draft and edit manuscript sections.
We have reviewed all AI-assisted work: code was tested, and every reported
number, table, and figure is regenerated by script from saved per-run results
and checked against them. We take responsibility for the final content of
this work, including text, claims, and artifacts produced with the aid of
generative AI.

\subsection*{Ethics statement}

Our experiments include health, credit, and recidivism data. Benchmark
accuracy and executable rules do not establish suitability for decisions
about individuals. MedCalc extraction uses supplied calculator identities
and schemas, and silver reference values; it is not a validated clinical
system. 
\subsection*{Reproducibility statement}

Section~\ref{sec:method} and Appendix~\ref{app:protocol} describe the learner,
information access, and evaluation protocol. Appendices~\ref{app:medcalc}
and~\ref{app:development} report extension and development results, while
Appendix~\ref{app:results} gives dataset-level tabular results. We release the codebase used for these experiments\ifdefined\ARXIV\ at
\url{https://github.com/Sylindril/CLIPPER}\fi.
\bibliography{references}
\bibliographystyle{iclr2027_conference}
\clearpage
\appendix
\section{Algorithms and evaluation details}
\label{app:protocol}

\subsection{Analysis of baselines}

In Figure~\ref{fig:all-baselines}, SLIPPER is our re-implementation, which was also used for the LLM harness.  The 
blue rule/tree
  baselines are Greedy\-Rule\-List, Tao\-Tree, FP\-Skope,
  Skope\-Rules, and imodels' own SLIPPER reimplementation, all from
  imodels \citep{singh2021imodels}, plus RIPPER (via wittgenstein); ensembles are
  Boosted\-Rules, Rule\-Fit, FIGS, and HS\-Tree from imodels.  CORELS
  is a separate implementation.
The datasets were adapted to be binary by making the positive class correspond to whatever category most directly corresponds to the more rare and/or interesting event, as follows:
college, ``Will a high-school student go to college?''; water potability, ``Will the water be safe for consumption?''; stroke, ``Will a patient suffer a stroke?', telco churn, ``Will a customer leave the company?'';
COMPAS, ``Will a defendant recidivate?''; HELOC, `` `Will a client repay within 2 years?''; adult, ``Will income exceed \$50K/yr?''; bank marketing.csv, ``Will a client subscribe?''

\subsection{Classification algorithm}

We use the following one-sided boosting procedure.
Let $a(p,q)=\frac12\log((p+\epsilon)/(q+\epsilon))$ and
$G(p,q)=\sqrt p-\sqrt q$, with $\epsilon=1/(2n)$.
The positive class is the rarer training class under the default class ordering.

\begin{enumerate}
\item Initialize $D(i)$ from normalized example weights. Compute total
positive and negative weights $(P,Q)$, set $b=a(P,Q)$, and update
$D(i)\propto D(i)\exp[-y_i b]$. Initialize the rule list to empty.
\item For each of the $J$ rounds, draw up to $M$ training examples without
replacement, with selection probabilities proportional to $D$. The sampled
wrappers retain relative weights, renormalized within the sample. $M=0$
means full data. MedCalc classification additionally balances the first
sample between classes where possible; subsequent samples use $D$ directly.
\item Propose a predicate. Discrete SLIPPER grows and prunes a conjunction
on this sample. CLIPPER receives its schema, labeled weighted rows, and the
configured statistics or tools, then returns a Python predicate.
\item Evaluate the proposed predicate on all training examples under $D$
to obtain $(p,q)$. Recompute total class weights $(P,Q)$.
If $G(p,q)>\max\{G(P,Q),G(Q,P)\}$, retain the rule with confidence
$a(p,q)$ and update covered weights by $\exp[-y_i a(p,q)]$.
Otherwise update the default confidence by $a(P,Q)$ and reweight every
example accordingly. Normalize $D$ after either update.
\item Return the default score and retained rules. Sum the confidence of
each matching rule and the default to predict. The implementation stores the
default in smoothed coverage coordinates, with a numerical nonnegativity
clamp when accumulating updates.
\end{enumerate}

A failed model call, malformed function, or construction-time execution error
can trigger up to three proposal attempts. If all fail, the classifier
builder returns an always-false predicate and the usual default comparison
applies. The tool-call limit is per attempt; it is not a bound on all calls
across a fit. Tool-enabled attempts have a request limit of the tool budget
plus five. The no-tools path has no iterative statistical-query interface.
The coding model's temperature and provider seed are unset, so a learner seed
does not fully determine its generated code.

\begin{table}[ht]
\caption{Configurations. All six conditions rescore
proposed rules on the full current training subset.}
\centering
\begin{tabular}{@{}ll r@{}}
\toprule
Configuration & Proposal sample & \\
\midrule
SLIPPER full & Full data  \\
SLIPPER M100 & $M=100$ \\
CLIPPER sample & $M=100$  \\
CLIPPER coverage & $M=100$  \\
CLIPPER hints & $M=100$ + hints  \\
CLIPPER hints+tools & $M=100$ + hints  \\
\bottomrule
\end{tabular}
\end{table}

\subsection{Prompts and statistical tools}

The classification system prompt is ``You are a concise, careful Python
programmer writing a single classification rule.'' The adopted task prompt
asks for exactly one conjunction, favors a large cluster with more positives
than negatives, and leaves other examples to subsequent rules. It requests
explicit handling of missing values and standard-library code. Labels and
weights are table columns, but are not keys available to the learned predicate.
The displayed sample weights are rescaled to have mean one.
The table format, full schema, quoted cell representation, and
\texttt{task\_variant\_2} framing are common to the main CLIPPER conditions.

The coverage tool evaluates a single Python Boolean expression on each
training row and returns the sums of current weights for covered positives
and negatives. It supplies aggregate statistics, not extra labeled rows.
The hints table supplies the best attribute test under discrete SLIPPER's
criterion, including both sample and full-data coverage. The hints tool accepts
a partial-rule filter and recomputes suggestions within that region.
We provide the full templates and tool docstrings in the supplementary source
files. Appendix~\ref{app:prompt-traces} gives representative prompt excerpts,
coverage-tool traces, and returned rules; full prompts accompany the source.

\subsection{Tabular data and preprocessing}

The suite builder uses a split seed of one, reserving a stratified 25\%
holdout for each task. Learning-curve seeds one through ten determine training
subsets drawn from the remaining rows. Each method uses the same subset for
a given dataset, size, and seed. The nine main sizes are
100, 175, 250, 400, 500, 600, 800, 1,200, and 2,000; a method may have only
the five-size grid in its archived runs. College supports sizes through 600.
The main figures fix the pool to the seven datasets that support 2,000.

Identifiers are removed from Stroke and Telco. Bank marketing excludes call
\texttt{duration}, which is known after the call. Adult excludes
\texttt{native-country}; COMPAS keeps nine demographic and criminal-history
features and excludes the COMPAS score columns. The current builder applies
this feature selection to the downloaded COMPAS rows; it does not implement
an additional ProPublica cohort filter. Missing entries remain null in
SLIPPER's input. Appendix tables use the prepared data, whose features can
differ from the source survey's summary counts.

\subsection{Text datasets}
\label{app:text-suite}

The text experiments (Figure~\ref{fig:text-results}b) use five two-class
datasets: SMS spam \citep{almeida2011sms}; Amazon, IMDb, and Yelp review
sentences labeled positive or negative \citep{kotzias2015group}; and
YouTube comment spam from five videos \citep{alberto2015tubespam}. Spam
makes up 13.4\% of SMS messages and 51.4\% of YouTube comments; the review
sets are balanced. As for tabular data, each dataset keeps a stratified
25\% test set, and every method trains on the same subset for each size
(50 to 750) and seed (one to ten).

CLIPPER sees each document's raw text, may use regular expressions, and
receives $M=100$ examples per proposal without coverage tools. SLIPPER
cannot read raw text, so it sees each document as its set of words and
searches all training rows. HS-Tree \citep{agarwal2022hierarchical} and the
other \texttt{imodels} baselines use TF-IDF word weights over at most 800
words. HS-Tree has the lowest error of these baselines from $n=200$; FIGS
is lower at $n=50$ and 100, by 0.8 and 0.2 points.
Table~\ref{tab:text-suite-error} gives test error for each dataset.

\begin{table}[t]
\caption{\textbf{Text classification, by dataset.} Test error (\%), mean
$\pm$ SEM over ten seeds; bold marks the lowest error at each size. Macro
rows average the five datasets.}
\label{tab:text-suite-error}
\centering
\footnotesize
\setlength{\tabcolsep}{2.5pt}
% Generated by source/plots/build_text_tables.py
\begin{tabular}{@{}lrrrrrrr@{}}
\toprule
Method & $n=50$ & $n=100$ & $n=200$ & $n=300$ & $n=400$ & $n=500$ & $n=750$ \\
\midrule
\multicolumn{8}{@{}l}{\textit{SMS spam}} \\
CLIPPER (no tools) & $\boldsymbol{8.4{\pm}1.5}$ & $\boldsymbol{4.9{\pm}1.0}$ & $\boldsymbol{3.5{\pm}0.5}$ & $\boldsymbol{2.9{\pm}0.4}$ & $\boldsymbol{2.2{\pm}0.1}$ & $\boldsymbol{2.6{\pm}0.2}$ & $\boldsymbol{2.0{\pm}0.3}$ \\
SLIPPER, full data & $11.9{\pm}0.6$ & $8.8{\pm}0.9$ & $7.1{\pm}0.4$ & $5.3{\pm}0.2$ & $5.5{\pm}0.4$ & $4.9{\pm}0.3$ & $3.7{\pm}0.2$ \\
HS-Tree & $12.5{\pm}0.5$ & $10.5{\pm}0.6$ & $8.2{\pm}0.3$ & $7.3{\pm}0.2$ & $7.3{\pm}0.4$ & $6.8{\pm}0.3$ & $6.1{\pm}0.3$ \\
\midrule
\multicolumn{8}{@{}l}{\textit{Amazon}} \\
CLIPPER (no tools) & $\boldsymbol{36.7{\pm}1.7}$ & $\boldsymbol{36.6{\pm}2.3}$ & $\boldsymbol{25.4{\pm}1.2}$ & $\boldsymbol{22.6{\pm}1.1}$ & $\boldsymbol{24.5{\pm}1.3}$ & $\boldsymbol{23.0{\pm}1.1}$ & $\boldsymbol{19.2{\pm}0.5}$ \\
SLIPPER, full data & $43.0{\pm}1.6$ & $43.8{\pm}1.4$ & $35.4{\pm}1.2$ & $30.9{\pm}0.9$ & $31.0{\pm}1.0$ & $32.6{\pm}1.1$ & $29.0{\pm}0.5$ \\
HS-Tree & $40.0{\pm}1.4$ & $41.4{\pm}1.7$ & $32.8{\pm}0.8$ & $30.7{\pm}0.8$ & $29.4{\pm}0.7$ & $28.6{\pm}0.5$ & $29.1{\pm}0.3$ \\
\midrule
\multicolumn{8}{@{}l}{\textit{IMDb}} \\
CLIPPER (no tools) & $\boldsymbol{40.7{\pm}1.2}$ & $\boldsymbol{37.0{\pm}1.6}$ & $\boldsymbol{33.7{\pm}2.0}$ & $\boldsymbol{30.1{\pm}0.7}$ & $\boldsymbol{28.1{\pm}0.7}$ & $\boldsymbol{27.4{\pm}0.7}$ & $\boldsymbol{24.8{\pm}0.5}$ \\
SLIPPER, full data & $44.9{\pm}1.3$ & $43.0{\pm}1.3$ & $41.5{\pm}1.2$ & $38.5{\pm}1.1$ & $37.0{\pm}0.9$ & $35.8{\pm}0.5$ & $34.7{\pm}0.3$ \\
HS-Tree & $45.0{\pm}1.0$ & $42.8{\pm}0.9$ & $41.2{\pm}0.9$ & $40.3{\pm}0.7$ & $38.6{\pm}0.8$ & $39.1{\pm}0.9$ & $37.2{\pm}0.0$ \\
\midrule
\multicolumn{8}{@{}l}{\textit{Yelp}} \\
CLIPPER (no tools) & $\boldsymbol{37.3{\pm}2.6}$ & $\boldsymbol{29.2{\pm}1.8}$ & $\boldsymbol{28.9{\pm}1.4}$ & $\boldsymbol{24.0{\pm}1.6}$ & $\boldsymbol{26.6{\pm}1.3}$ & $\boldsymbol{22.5{\pm}1.0}$ & $\boldsymbol{19.6{\pm}0.2}$ \\
SLIPPER, full data & $44.6{\pm}1.5$ & $40.8{\pm}1.0$ & $34.4{\pm}1.7$ & $34.1{\pm}1.1$ & $31.2{\pm}1.1$ & $30.2{\pm}0.6$ & $28.8{\pm}0.3$ \\
HS-Tree & $45.7{\pm}1.6$ & $38.2{\pm}1.4$ & $32.1{\pm}0.6$ & $32.2{\pm}0.8$ & $28.7{\pm}0.7$ & $29.1{\pm}0.5$ & $27.4{\pm}0.2$ \\
\midrule
\multicolumn{8}{@{}l}{\textit{YouTube spam}} \\
CLIPPER (no tools) & $18.1{\pm}4.2$ & $\boldsymbol{8.7{\pm}0.7}$ & $\boldsymbol{8.4{\pm}0.5}$ & $\boldsymbol{9.1{\pm}0.8}$ & $\boldsymbol{8.4{\pm}0.7}$ & $8.3{\pm}0.5$ & $8.3{\pm}0.4$ \\
SLIPPER, full data & $26.7{\pm}4.7$ & $23.4{\pm}4.8$ & $22.5{\pm}4.5$ & $25.0{\pm}4.7$ & $24.7{\pm}4.1$ & $29.4{\pm}3.6$ & $34.0{\pm}0.3$ \\
HS-Tree & $\boldsymbol{18.0{\pm}2.5}$ & $12.3{\pm}0.5$ & $10.1{\pm}0.5$ & $9.7{\pm}0.3$ & $9.0{\pm}0.5$ & $\boldsymbol{8.2{\pm}0.4}$ & $\boldsymbol{6.6{\pm}0.3}$ \\
\midrule
\multicolumn{8}{@{}l}{\textit{Macro}} \\
CLIPPER (no tools) & $\boldsymbol{28.2{\pm}1.1}$ & $\boldsymbol{23.3{\pm}0.7}$ & $\boldsymbol{20.0{\pm}0.6}$ & $\boldsymbol{17.7{\pm}0.4}$ & $\boldsymbol{18.0{\pm}0.4}$ & $\boldsymbol{16.8{\pm}0.3}$ & $\boldsymbol{14.8{\pm}0.2}$ \\
SLIPPER, full data & $34.2{\pm}1.1$ & $32.0{\pm}1.1$ & $28.2{\pm}1.0$ & $26.8{\pm}1.0$ & $25.9{\pm}0.9$ & $26.6{\pm}0.8$ & $26.0{\pm}0.2$ \\
HS-Tree & $32.2{\pm}0.7$ & $29.1{\pm}0.5$ & $24.9{\pm}0.3$ & $24.1{\pm}0.3$ & $22.6{\pm}0.3$ & $22.4{\pm}0.2$ & $21.3{\pm}0.1$ \\
\bottomrule
\end{tabular}

\end{table}

\textbf{A dataset published after the model's training cutoff.} The five
datasets above are long-public benchmarks that the model may have seen during
training. We therefore also use Inflation Research Abstracts
\citep{gonzalez2025inflation}, which was added to the UCI repository in
February 2025, after the training cutoff of Gemini~3.1 Flash-Lite. It
contains 1,138 abstracts of economics papers on inflation, labeled by whether
the paper uses machine-learning or AI methods (20.9\% positive). The abstracts
themselves are older, but their labels were created for a 2025 study. We split
the data and run every method as for the five datasets above, with 853
training and 285 test abstracts. CLIPPER reading the raw text has the lowest
error at every training size, 6.9\% at $n=750$ versus 8.4\% for full-data
SLIPPER and 9.0\% for CLIPPER reading word sets
(Table~\ref{tab:text-post-cutoff}).

\begin{table}[t]
\caption{\textbf{Inflation Research Abstracts, published after the
model's training cutoff.} Test error (\%), mean $\pm$ SEM over ten seeds;
bold marks the lowest error at each size. CLIPPER uses no coverage tools and
$M=100$. The \texttt{imodels} baselines use TF-IDF word weights, except CORELS,
which uses word-presence indicators.}
\label{tab:text-post-cutoff}
\centering
\footnotesize
\setlength{\tabcolsep}{2.5pt}
% Generated by source/plots/build_text_tables.py
\begin{tabular}{@{}lrrrrrrr@{}}
\toprule
Method & $n=50$ & $n=100$ & $n=200$ & $n=300$ & $n=400$ & $n=500$ & $n=750$ \\
\midrule
CLIPPER, raw text & $\boldsymbol{13.2{\pm}1.7}$ & $\boldsymbol{9.6{\pm}0.7}$ & $\boldsymbol{8.5{\pm}0.4}$ & $\boldsymbol{8.4{\pm}0.4}$ & $\boldsymbol{7.4{\pm}0.3}$ & $\boldsymbol{7.8{\pm}0.5}$ & $\boldsymbol{6.9{\pm}0.2}$ \\
CLIPPER, word sets & $13.6{\pm}1.7$ & $9.7{\pm}0.4$ & $9.7{\pm}0.6$ & $9.2{\pm}0.2$ & $8.7{\pm}0.4$ & $10.0{\pm}0.5$ & $9.0{\pm}0.3$ \\
SLIPPER, full data & $17.7{\pm}1.5$ & $14.0{\pm}0.8$ & $10.9{\pm}0.5$ & $10.6{\pm}0.5$ & $9.8{\pm}0.2$ & $9.5{\pm}0.3$ & $8.4{\pm}0.3$ \\
\midrule
HS-Tree & $23.1{\pm}1.9$ & $18.1{\pm}1.3$ & $15.8{\pm}1.2$ & $16.2{\pm}0.9$ & $14.4{\pm}0.5$ & $13.4{\pm}0.8$ & $13.4{\pm}0.4$ \\
Boosted stumps & $20.2{\pm}1.2$ & $16.7{\pm}1.1$ & $12.5{\pm}0.5$ & $11.8{\pm}0.4$ & $11.5{\pm}0.4$ & $11.6{\pm}0.2$ & $11.3{\pm}0.3$ \\
CORELS & $16.0{\pm}1.3$ & $14.1{\pm}0.3$ & $14.1{\pm}0.3$ & $14.6{\pm}0.3$ & $13.5{\pm}0.3$ & $13.7{\pm}0.3$ & $13.3{\pm}0.2$ \\
FIGS & $21.8{\pm}2.1$ & $19.1{\pm}1.8$ & $16.0{\pm}1.4$ & $13.6{\pm}0.9$ & $13.0{\pm}0.4$ & $12.1{\pm}0.7$ & $12.3{\pm}0.4$ \\
TAO tree & $20.0{\pm}2.2$ & $19.2{\pm}1.9$ & $16.4{\pm}1.3$ & $14.2{\pm}0.7$ & $12.6{\pm}0.4$ & $13.3{\pm}0.7$ & $12.2{\pm}0.4$ \\
\bottomrule
\end{tabular}

\end{table}

\subsection{Comparison with RuleChef on BANKING77}
\label{app:banking77}

Recent systems also use an LLM to write rules. The closest to CLIPPER,
RuleChef \citep{kovacs2026rulechef}, asks an LLM to write rules from labeled
examples, but without boosting: it writes a small set of rules and then
revises them to fix their mistakes. We compare the two on the benchmark
RuleChef's authors use: five intents from BANKING77
\citep{casanueva2020efficient}. Both methods read the raw text and use the
same model, Gemini~3.1 Flash-Lite, and RuleChef runs its authors' own
benchmark script and settings.

CLIPPER's error is lower than RuleChef's in all 50 runs (five training sizes,
ten seeds each), by 29.7 points on average with coverage tools and 27.8
points without them (Figure~\ref{fig:text-results}c and
Table~\ref{tab:banking77-appendix}). From $n=100$ on, RuleChef's error stays between
34.9 and 39.7\%, while CLIPPER with tools reaches 8.0\% at $n=200$.
RuleChef's rules label only about half of the test queries: after its first
rounds of revision, it usually replaces its initial rules with one broad
regular expression per intent.

The five intents are beneficiary not allowed, card arrival, disposable card
limits, exchange rate, and pending cash withdrawal. Training queries come
from BANKING77's official training split (685 queries), and the test set is
all 200 official test queries for these intents, 40 each. For each size (50,
100, 200, 400, and 600) and seed (one to ten), both methods train on the same
random subset.

CLIPPER reads each query's raw text, may use regular expressions, and
receives $M=100$ examples per proposal. We run it with the coverage tools used
for MedCalc (Appendix~\ref{app:medcalc}), at most six calls per proposal, and
without them. As for MedCalc, it fits one two-class model per intent for six
rounds and predicts the intent with the highest centered score.

We run RuleChef 0.2.2 through its authors' BANKING77 script, unchanged, with
the settings of their published run: regular-expression rules, at most 100
rules, 200 examples shown to the model, up to 15 rounds of revision, and
their agentic coordinator. The script writes rules from five examples per
intent and revises them against the other $n-25$ training queries. When no
rule matches a test query, RuleChef returns no label; we assign the most
common intent in that training subset, as SLIPPER's default rule does.
Counting these queries as errors instead raises RuleChef's error to 47 to
66\%.

\begin{table}[t]
\caption{\textbf{BANKING77.} Test error (\%), mean $\pm$ SEM over ten
seeds; bold marks the lowest error at each size. When no RuleChef rule
matches a query, it receives the most common intent in the training subset.}
\label{tab:banking77-appendix}
\centering
\small
% Generated by source/plots/build_banking77.py
\begin{tabular}{@{}lrrrrr@{}}
\toprule
Method & $n=50$ & $n=100$ & $n=200$ & $n=400$ & $n=600$ \\
\midrule
CLIPPER (tools) & $\boldsymbol{14.6{\pm}0.9}$ & $\boldsymbol{14.1{\pm}0.6}$ & $\boldsymbol{8.0{\pm}0.6}$ & $\boldsymbol{8.2{\pm}0.8}$ & $\boldsymbol{8.4{\pm}0.4}$ \\
CLIPPER (no tools) & $16.8{\pm}1.1$ & $16.0{\pm}1.1$ & $11.0{\pm}0.5$ & $10.4{\pm}0.6$ & $9.1{\pm}0.6$ \\
RuleChef & $53.7{\pm}3.0$ & $36.1{\pm}3.2$ & $39.7{\pm}3.9$ & $37.7{\pm}2.5$ & $34.9{\pm}1.7$ \\
\bottomrule
\end{tabular}

\end{table}

\subsection{Baseline settings and model size}

We use \texttt{imodels} version 2.0.4 for the specialist baselines.
Continuous inputs are median-imputed and categorical inputs one-hot encoded;
preprocessing fits on the current training subset only. Binary-input learners
such as CORELS additionally discretize continuous values into four quantile
bins learned on that subset. Unknown test categories are ignored by the encoder.
We use estimator defaults. RuleFit enables its own internal selection, and
hierarchical shrinkage trees select shrinkage by internal cross-validation.
We set a 600-second timeout per baseline fit.

\begin{table}[ht]
\caption{Baseline settings. These controls
constrain different objects and do not define a common complexity budget.}
\centering
\begin{tabular}{@{}lp{0.70\linewidth}@{}}
\toprule
Method & Settings \\
\midrule
RuleFit & 100 source trees, tree size 4, maximum rules 30,
linear terms included, internal cross-validation enabled \\
FIGS & Maximum 12 splits \\
Boosted stumps & 15 estimators, depth one \\
Greedy rule list & Maximum depth 5 \\
SkopeRules & 10 estimators, depth 3, minimum precision .5 and recall .01 \\
CORELS & Regularization .01, maximum antecedent cardinality 2,
minimum support .01, 10,000 search iterations \\
\bottomrule
\end{tabular}
\end{table}

A curve seed is passed to an estimator only when its constructor exposes
\texttt{random\_state}; all seeds still identify shared training subsets.
The separate \texttt{imodels} SLIPPER implementation failed in 50 of 380
dataset/size/seed cells, including every Water potability cell, and we exclude it from
the common aggregate. The failure is reported by its AdaBoost wrapper when
the base classifier is worse than random.

At $n=2{,}000$, the mean numbers of retained predicates on the seven datasets
are 8.9 for full SLIPPER, 9.6 for CLIPPER sample, 11.8 for CLIPPER
coverage, 16.0 for CLIPPER hints, and 17.4 for hints+tools. Each Python
predicate can itself contain several operations, so these counts describe
ensemble size rather than human effort. 
\subsection{Alternative harnesses}

Covering removes covered examples and redistributes weight uniformly over
survivors. Its positive-disjunction form accepts a rule only if $p>q$,
and stops after one failed proposal unless a larger patience is specified.
The discrete legacy condition combines FOIL-style growth with holdout pruning;
its growth criterion differs from this acceptance gate.

One-shot is one outer round of this covering procedure. The LLM receives
$M=100$ examples and is asked for a complete disjunctive classifier inside
one function. The cross-validated round budget becomes an approximate
branch-count hint in its prompt. The entire function must pass the same
$p>q$ gate, so a rejected proposal leaves a default-only classifier. The
one-shot rule count indicates whether that function was retained;
it does not count its internal branches. This changes hypothesis composition,
acceptance, and total compute together with iteration.

\subsection{Uncertainty and repetitions}

Macro error averages the $D$ dataset means equally, and its SEM is
$\sqrt{\sum_d(\mathrm{SEM}_d/D)^2}$. SEM summarizes variation across the ten seeded training subsets and fits on
a fixed holdout. It is not uncertainty over newly sampled datasets or new
test sets. In particular, sizes from the same dataset are dependent and must
not be treated as independent experiments in a pooled significance test.
We compare observed means. 
For both MedCalc tasks, SEM measures variation over ten learner seeds with
fixed training and evaluation rows. We compute calculator accuracy and
extraction F1 separately for each seed, then average the ten scores.
Resampling evaluation notes would estimate uncertainty conditional on each
fitted model. The deterministic pool has no training-seed
variation.

\clearpage
\section{MedCalc protocols and additional results}
\label{app:medcalc}

\subsection{Data and calculator prediction}

We use \texttt{nsk7153/MedCalc-Bench-Verified} at revision
\path{591157b3343b4dda247294f9d929da4c75026fa8}, derived from
MedCalc-Bench \citep{khandekar2024medcalc}. Ten rows per calculator are
held out from training before applying a training cap. The 55-way official
test split has 1,100 rows, 20 per calculator. The realized training sizes at caps
1, 4, 16, and 64 are 55, 220, 870, and 3,302. Each method uses the same
training rows at a given cap, with ten learner seeds.

SLIPPER and the other discrete learners cannot search over free text, so
we represent each note by the set of words it contains. We lowercase the note,
split it into maximal runs of the letters a--z, drop one-letter tokens, and
remove duplicates. The vocabulary is built from the training rows alone: we
keep words that appear in at least five training notes, up to the 5,000 most
frequent, and remove all other words from training and test
notes. A discrete rule condition then tests whether a given word is present,
and a rule is a conjunction of such tests, such as ``the note contains
\texttt{wells} and \texttt{dvt}.'' This representation discards numbers,
punctuation, word order, word counts, and therefore negation and phrases; a
rule can test whether a note mentions \texttt{creatinine} but not its value.
CLIPPER instead receives the raw note as the text attribute \texttt{NOTE} and
can use its full content, including numbers and phrases. Neither method
receives the question.
The one-versus-rest runner uses six rounds and first-round stratified sampling,
without internal round selection. A class prediction is the argmax of the
55 signed scores after removing their final learned default intercepts;
lowest calculator ID breaks ties. Exact 55-way accuracy is primary.
Calculator-family merging changes the prediction target. A sample cap limits
rule proposal, while all retained rules receive
full-training-set scoring and the default CLIPPER condition has coverage tools.
We omit the optimization-criterion wording from the prompts for both
calculator prediction and extraction.

\begin{figure}[ht]
\centering
\includegraphics[width=0.6\linewidth]{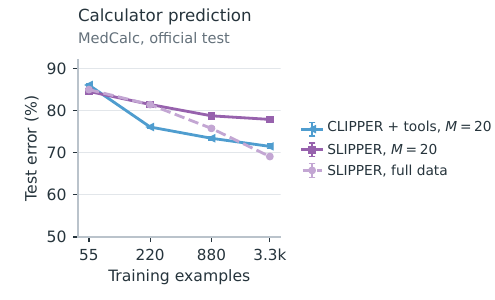}
\caption{\textbf{Calculator prediction on MedCalc.} Test error (100 minus
exact 55-way accuracy) on the official test split, mean and SEM over ten
seeds. CLIPPER reads the raw notes; SLIPPER reads their word sets.}
\label{fig:medcalc-calculators}
\end{figure}

Reading the raw notes, CLIPPER predicts the calculator more accurately than
SLIPPER given the same sample of 20 examples per proposal at every training
size from 220 notes on, and more accurately than full-data SLIPPER at 220 and
880 notes (Figure~\ref{fig:medcalc-calculators}).
Table~\ref{tab:medcalc-final} gives the exact accuracies, including $M=100$.

\begin{table}[ht]
\caption{Calculator prediction on the balanced official test. Exact
accuracy (\%), mean $\pm$ SEM over ten seeds per configuration and cap.}
\label{tab:medcalc-final}
\centering
\begin{tabular}{@{}rrrrrr@{}}
\toprule
Cap & \multicolumn{3}{c}{SLIPPER} & \multicolumn{2}{c}{CLIPPER (raw notes)} \\
\cmidrule(lr){2-4}\cmidrule(l){5-6}
 & Full & $M=20$ & $M=100$ & $M=20$ & $M=100$ \\
\midrule
1 & $15.0\pm0.3$ & $15.5\pm0.5$ & $15.0\pm0.3$ & $13.9\pm0.3$ & $10.1\pm0.5$ \\
4 & $18.6\pm0.3$ & $18.6\pm0.3$ & $20.1\pm0.4$ & $23.9\pm0.3$ & $22.1\pm0.3$ \\
16 & $24.2\pm0.3$ & $21.2\pm0.3$ & $25.8\pm0.5$ & $26.6\pm0.4$ & $25.9\pm0.3$ \\
64 & $30.9\pm0.4$ & $22.1\pm0.4$ & $25.9\pm0.4$ & $28.5\pm0.6$ & $28.2\pm0.4$ \\
\bottomrule
\end{tabular}

\end{table}

\subsection{Value extraction}

Extraction is conditioned on the true calculator and its requested fields.
We keep the 33 calculators whose entity-key sets are consistent across their
training and test rows, giving 34 distinct keys. Inconsistent omitted-versus-false
annotations make the excluded schemas unsuitable for this exact-value task.
The reference labels are silver \texttt{Relevant Entities} annotations.
We select the ten most frequent keys after capping training at 150 rows per
calculator, breaking ties alphabetically: sodium, weight, height, sex,
bicarbonate, chloride, age, glucose, heart rate or pulse, and QT interval.
These span 25 calculators. The official test split contains 500 instances
and 1,120 requested values.

We use ten learner seeds on fixed training rows.
Appendix~\ref{app:extraction-algorithm} gives the training and prediction procedure.

Canonicalization folds number formatting, case, spacing, and unit spelling.
It does not perform unit conversion or general synonym matching. Precision,
recall, and micro F1 compare predicted and gold sets of key/value pairs.
An incorrect returned value contributes both a false positive and a missed
gold value; abstention contributes only a missed gold value. Complete-row
accuracy requires exact equality of all selected values for that row.

We wrote the hand-written regex pools using Anthropic's Claude Fable 5.1
as our coding agent. We inspected the training examples, checked the patterns'
precision and coverage on them, and selected rules from the resulting pool
using the same scoring procedure. All ten selected keys have a pool.
For all three approaches, we train each model once using only the training
data, then evaluate it on the official test split.

\begin{table}[ht]
\caption{Extraction official-test results (\%), mean $\pm$ SEM over ten seeds.
The deterministic pool is evaluated once. ``All selected'' is complete-row accuracy.}
\label{tab:extraction-final}
\centering
\small
\setlength{\tabcolsep}{4pt}
\begin{tabular}{@{}lrrrr@{}}
\toprule
Rule source & Precision & Recall & F1 & All selected \\
\midrule
Hand-written pool & $97.0$ & $93.7$ & $95.3$ & $91.0$ \\
Python, $M=20$ & $95.6\pm0.3$ & $90.4\pm0.6$ & $92.9\pm0.3$ & $84.7\pm1.3$ \\
Python, $M=55$ & $96.1\pm0.2$ & $91.4\pm0.5$ & $93.7\pm0.2$ & $87.0\pm0.8$ \\
Python, $M=110$ & $96.1\pm0.2$ & $90.6\pm0.5$ & $93.3\pm0.3$ & $84.7\pm1.1$ \\
Regex, $M=20$ & $96.3\pm0.2$ & $86.4\pm1.1$ & $91.1\pm0.6$ & $77.7\pm2.0$ \\
Regex, $M=55$ & $96.3\pm0.1$ & $86.0\pm0.6$ & $90.8\pm0.3$ & $77.4\pm1.2$ \\
Regex, $M=110$ & $96.2\pm0.2$ & $85.8\pm1.2$ & $90.7\pm0.7$ & $77.2\pm2.1$ \\
\bottomrule
\end{tabular}

\end{table}

On official test, the pool has 1,049 true positives, 33 false positives, and
71 false negatives, with all selected fields correct on 455 of 500 rows.

\subsection{Extraction with SLIPPER and CLIPPER}
\label{app:extraction-algorithm}

Each training example contains \texttt{NOTE}, the supplied \texttt{CALC\_ID},
and the requested \texttt{ENTITY}; its target is that entity's annotated
value. We pool examples of the same entity across calculators and learn
one ensemble per entity key. SLIPPER supplies the boosting procedure:
sampling examples, scoring rules, assigning confidences, and updating
weights. The hand-written baseline selects rules from a fixed regex pool;
CLIPPER generates a new candidate with the coding model.

For a key with $n$ training examples, initialize $D_1(i)=1/n$ and an empty
ensemble. The default prediction is abstention. Use a budget of ten rounds:
\begin{enumerate}
\item \textbf{Propose a rule.} CLIPPER draws up to $M\in\{20,55,110\}$
examples without replacement using $D_t$. The prompt contains their inputs,
gold values, weights, and schema. Gemini 3.1 Flash-Lite returns either a
Python function \texttt{extract(x)} or a regex pattern, flags, and output
template. Both return a value or \texttt{None}. Python can perform arithmetic;
the regex template combines captured text and literals. Neither arm uses
coverage tools. The pool baseline instead considers every hand-written rule.

\item \textbf{Score on all training examples.} For a candidate $r_t$, let
$s_t(i)$ be $+1$ for a correct returned value, $-1$ for an incorrect return,
and $0$ for abstention, comparing values after canonicalization. Compute
\begin{equation}
 p_t=\sum_{i:s_t(i)=1}D_t(i),\qquad
 q_t=\sum_{i:s_t(i)=-1}D_t(i).
\end{equation}
The pool baseline chooses the candidate with the largest
$1+\sqrt{p_t}-\sqrt{q_t}$. CLIPPER's candidate is scored by the same procedure,
including on training examples outside its prompt sample.

\item \textbf{Retain and reweight.} If $p_t>q_t$, retain the rule with
confidence $\alpha_t$ and update
\begin{equation}
 \alpha_t=\frac12\log\frac{p_t+\epsilon}{q_t+\epsilon},\qquad
 D_{t+1}(i)=\frac{D_t(i)e^{-\alpha_t s_t(i)}}{Z_t},\qquad
 \epsilon=\frac{1}{2n},
\end{equation}
where $Z_t$ normalizes the weights. Before normalization, correct returns
multiply an example's weight by $e^{-\alpha_t}$, incorrect returns by
$e^{\alpha_t}$, and abstentions leave it unchanged.
If $p_t\le q_t$, CLIPPER discards the candidate
and continues with unchanged weights. The pool baseline stops, since its
best candidate failed the same test.
\end{enumerate}

A failed CLIPPER proposal has up to two retries; exhausting them adds no rule that
round. Execution errors and one-second timeouts count as abstentions.
Duplicate retained rules are merged by adding their confidences.

At prediction time, run the saved rules for each requested key. Group their
outputs by canonical value, sum the confidences within each group, and return
the value with the largest sum. If no rule returns a value, abstain. This
step uses no LLM calls. We apply the fitted ensembles to the official test
split, then join the key predictions back into each calculator row
for scoring.

\subsection{Example extraction prompts and rules}
\label{app:rule-example}

These excerpts show CLIPPER's extraction prompts and three generated rules.
We show only the relevant text and fields from each sampled example.

\begingroup
\lstset{basicstyle=\ttfamily\scriptsize,breaklines=true,breakatwhitespace=true,
  columns=fullflexible,keepspaces=true,showstringspaces=false,
  upquote=true,aboveskip=4pt,belowskip=4pt,breakindent=12pt,
  escapeinside={(*@}{@*)}}

% Regex prompts and rules: ../slipper/aditya_task_dump/
% medcalc_repeats_20260915_v3/cells/extraction/extract_regex_M20/seed01/
% age/attempt_001/ (round 0) and sex/attempt_001/ (round 1).
\paragraph{Captures and literals.}
The regex prompt specifies how the learner will execute the returned rule:
\begin{lstlisting}
Task: write a rule as a REGEX SPEC -- structured fields, not code.  The learner
runs it as
    m = re.search(pattern, x['NOTE'], flags)
and builds the value from `template`, so fill in these three fields:
\end{lstlisting}
Here \texttt{i} ignores case and \texttt{1} inserts the first captured group.
For age, a sampled example and the returned rule are:
\begin{lstlisting}
ENTITY: age
NOTE: "An 89-year-old Japanese woman..."
GOLD: [89, "years"]

pattern: (\d+(?:\.\d+)?)-year-old
flags: i
template: [1, "years"]
\end{lstlisting}
The capture supplies the number; the template supplies the unit.
For sex, another prompt and returned rule illustrate a literal output:
\begin{lstlisting}
ENTITY: sex
NOTE: "A 31-year-old woman..."
GOLD: "Female"

pattern: \b(?:woman|female|girl)\b
flags: i
template: ["Female"]
\end{lstlisting}
This rule returns \texttt{Female} for any of the three matched words and
abstains when none occurs.

\paragraph{Python and arithmetic.}
The Python prompt asks for a value-returning function:
\begin{lstlisting}
Task: write a rule in the form of a Python function
    def extract(x: dict): ...
that returns the VALUE asked for by `x['ENTITY']`, read out of the note text in
`x['NOTE']`, for a meaningful cluster of examples -- and returns None on every
other example.
\end{lstlisting}
% Python prompt: source/prompts/templates.json, _TASK_DEFINITION_EXTRACT_PYTHON.
% Platelet rule and note: ../slipper/aditya_task_dump/runs/
% extract_llm_python_final/platelet_count/{platelet_count.shyp,platelet_count.json}.
A platelet-count rule from the earlier 34-key experiment includes the
following branch, with \texttt{re} already imported:
\begin{lstlisting}
note = x.get('NOTE')
if not note or x.get('ENTITY') != 'Platelet count':
    return None
m = re.search(
    r'platelet(?: count)?(?: of)?\s+(\d+(?:\.\d+)?)\s*k[/\s]?(?:(*@\ensuremath{\mu}@*)|u)l',
    note, re.IGNORECASE)
if m:
    return [float(m.group(1)) * 1000, '(*@\ensuremath{\mu}@*)L']
\end{lstlisting}
For the training-note excerpt below, the branch returns the annotated value:
\begin{lstlisting}
NOTE: "...platelet count of 233 k/(*@\ensuremath{\mu}@*)L..."
GOLD / RETURNED VALUE: [233000.0, "(*@\ensuremath{\mu}@*)L"]
\end{lstlisting}
Python expresses the multiplication by 1,000; a capture template can only
combine matched text and literals.
\endgroup

\clearpage
\section{Development experiments and covering controls}
\label{app:development}

\subsection{Development suite and configuration selection}

Our development suite contains Abalone, Breast cancer Wisconsin, Cylinder
bands, German credit, Hypothyroid, Move1, Obesity, Pima diabetes, Portuguese
student performance, and Titanic. It was selected for binary targets,
meaningful attribute names, and varied domains. Nine datasets use a fixed,
stratified 25\% holdout with split seed one; Move1 retains its native split.
Prompt and configuration comparisons use ten training seeds. The holdouts in
this suite guide development and are distinct from the tabular evaluation
suite in the main text.

Abalone labels age through a threshold of ten rings. Obesity collapses the
three obesity categories and removes height and weight. Student performance
uses final grade below ten and excludes the two preceding grades. German
credit codes are decoded to meaningful attribute values. Missing data remain
explicit; the Pima source's zero-valued measurements were not converted to
null. The Obesity source includes synthetic augmentation. These choices are
part of the task definitions we evaluate.

\begin{table}[ht]
\caption{Development configurations in the full-data comparison.
Macro test error (\%) averages ten dataset means, with ten seeds per dataset.
Rows are different configurations, not successive additions to one system.}
\centering
\begin{tabular}{@{}lr@{}}
\toprule
Configuration & Macro error \\
\midrule
Discrete SLIPPER, full data & 18.1 \\
CLIPPER, table sample, no tools (earlier prompt) & 20.0 \\
Per-attribute SLIPPER condition tools & 19.2 \\
One-conjunction framing with coverage tool & 19.0 \\
Revised coverage-tool prompt & 18.8 \\
Condition-hints table with iterative queries & \textbf{17.7} \\
\bottomrule
\end{tabular}
\end{table}

\paragraph{Sample size.}
In the dedicated sample-size comparison, increasing $M$ from 100 to 200 and
400 gives macro errors of 20.5, 20.8, and 20.9\%. These values come from
a different configuration comparison than the table above. The larger prompt
samples do not improve average error. Coverage-retrieval probes additionally
record counting errors on long samples, motivating the coverage tools. The
sample-size curve alone does not identify the cause of the degradation.

\paragraph{Prompt representation and task framing.}
Replacing per-example dictionaries with a table reduced prompt tokens by
about 71\%, with a 0.5-point macro-error increase. We retained the full
schema, quoted cell values, and table format in our main configuration.
Most wording changes had small measured effects.
The one-conjunction-per-round task framing improved macro error by
1.0 points relative to its prompt control. Its instruction is still a
prompt preference, rather than a syntactic check on generated predicates.

\paragraph{Statistical feedback.}
Per-attribute optimizers expose the best full-data threshold or category.
The condition table packages these suggestions with sample and full-data
coverage; its tool form recomputes them within a partial rule's coverage.
The latter achieved the lowest development macro in the table, 17.7\%.
The separate initial-table comparison gave 18.8\% and varied across datasets.
We therefore evaluate initial hints and iterative queries as separate
feedback conditions in the main experiments.

\subsection{Repairs to discrete covering}
\label{app:setcover}

The legacy covering learner uses a positive disjunction and accepts rules
only when their residual positive coverage exceeds negative coverage. Two
repairs change how discrete search reaches that gate. Gate-aware sizing
retains only pruned generalizations that still pass the gate and prefers the
more specific rule on ties. Linear growth replaces the FOIL-style criterion
with $p-q+1$, matching the change in covered training error. Two further
variants learn ordered decision lists with per-rule consequents; these
change the hypothesis class as well as the search.

The six conditions each contain 440 dataset/size/seed cells: ten development
datasets at sizes 100, 200, 300, and 400, plus the four datasets that support
size 800, with ten seeds in every group. The pooled cell mean below weights
these 440 cells equally. It is not the seven-dataset test macro used in the
main tables.

\begin{table}[ht]
\caption{Completed ten-seed discrete covering comparisons on the development
suite. Cells with no retained rule predict the fallback class.}
\centering
\begin{tabular}{@{}lrr@{}}
\toprule
Condition & Pooled error (\%) & Zero-rule cells / 440 \\
\midrule
Legacy covering & 23.8 & 50 \\
Gate-aware sizing & 24.1 & 11 \\
Linear growth & 23.3 & 31 \\
Gate-aware sizing + linear growth & \textbf{23.0} & 1 \\
Decision list, FOIL growth & 23.9 & 20 \\
Decision list, linear growth & 23.5 & 71 \\
\bottomrule
\end{tabular}
\end{table}

The combined repair improves on legacy covering in 206 cells, worsens in
186, and ties in 48. It reduces empty fits substantially, but at size 800
its pooled error is 22.5\%, compared with 21.2\% for legacy covering.
Thus the repair's effect depends on training size. On the seven main test
datasets, ten seeds give $24.6\pm0.2$\% error at $n=250$ and
$22.8\pm0.1$\% at $n=1{,}200$ (Table~\ref{tab:covering-full}), using the
same training subsets and training-only cross-validated round budgets.

\subsection{Covering patience with the LLM proposer}

Patience three skips a failed proposal and permits a new proposal under the
same residual distribution; patience one stops immediately. Both conditions
have 380 cells on the tabular evaluation suite, covering the five-size grid
100, 250, 600, 1,200, and 2,000 with ten seeds per feasible dataset and size.
Their pooled cell errors are 23.0\% and 22.9\% for patience one and three.
Zero-rule fits fall from 76 to 56, while total model-call cost rises from
\$22.93 to \$44.11. These pooled summaries have a changing dataset pool;
Table~\ref{tab:covering-full} recomputes the size-specific comparisons on the
fixed seven-dataset pool. The results show more retained rules with mixed
accuracy effects, and qualify a blanket claim that boosting outperforms
covering for an LLM rule proposer.

\begin{table}[ht]
\caption{All outer-loop conditions on the seven main tabular datasets.
Error (\%), mean $\pm$ propagated SEM over ten seeds. Discrete search uses
full training data; the LLM proposer uses $M=100$. Patience~3 permits up to
two further proposals after a rejected rule. The last row uses gate-aware
pruning and linear growth.}
\label{tab:covering-full}
\centering
\begin{tabular}{@{}lrr@{}}
\toprule
Harness / rule proposer & $n=250$ & $n=1{,}200$ \\
\midrule
Boosting / SLIPPER & $23.9\pm0.2$ & $\boldsymbol{22.0\pm0.1}$ \\
Legacy covering / SLIPPER & $25.7\pm0.3$ & $23.9\pm0.2$ \\
Boosting / CLIPPER coverage & $23.2\pm0.2$ & $22.4\pm0.1$ \\
Covering / CLIPPER, patience 1 & $\boldsymbol{23.0\pm0.1}$ & $22.5\pm0.1$ \\
Covering / CLIPPER, patience 3 & $23.5\pm0.2$ & $22.2\pm0.1$ \\
One-shot / CLIPPER coverage & $23.9\pm0.3$ & $24.8\pm0.4$ \\
One-shot / CLIPPER hints & $26.6\pm0.4$ & $26.8\pm0.3$ \\
\midrule
Repaired covering / SLIPPER & $24.6\pm0.2$ & $22.8\pm0.1$ \\
\bottomrule
\end{tabular}

\end{table}

\clearpage
\section{Dataset-level tabular results}
\label{app:results}
\begin{figure}[ht]
\centering
\includegraphics[width=\linewidth]{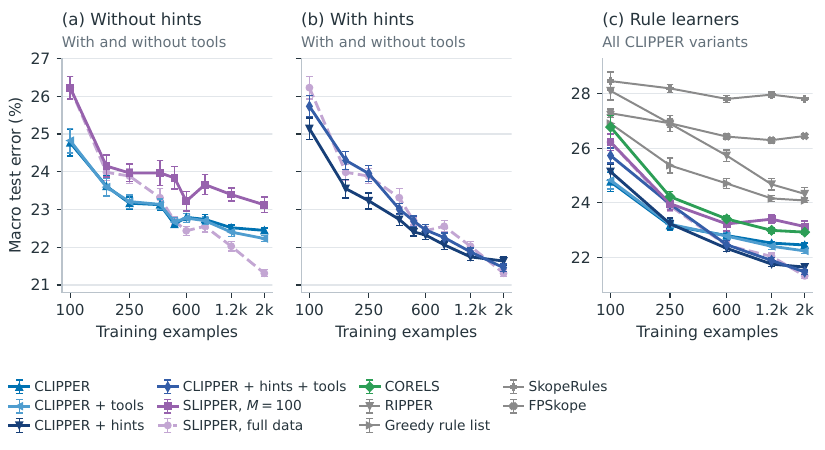}
\caption{\textbf{All CLIPPER variants on tabular data.} The curves of
Figure~\ref{fig:tabular-results} for every CLIPPER variant: (a) without
hints and (b) with hints, each with and without coverage tools, and (c) all
four variants against established rule learners. Macro test error on seven
tabular datasets (mean and propagated SEM, ten seeds).}
\label{fig:tabular-all}
\end{figure}

We plot the fixed seven-dataset macro used in the main figures and tables for
every CLIPPER variant (Figure~\ref{fig:tabular-all}), followed by
dataset-level results that include College where available.

Each cell is test-error percentage, mean $\pm$ SEM over seeds.
Macro error is the unweighted mean across datasets; its SEM is
$\sqrt{\sum_d(\mathrm{SEM}_d/D)^2}$.
In the dataset-level tables, bold values mark the lowest mean in each row
among the methods shown in that panel, including ties. Panels showing only
one method are not bolded. Bold does not indicate statistical significance.
We compute mean rank across all methods in each comparison, including
methods shown in separate panels.

\subsection{Discrete baselines}
\label{app:symbolic}
% Imported from source/tables/hero-250-symbolic.md
\begin{longtable}{@{}lrrr@{}}
\caption{Discrete baselines at $n=250$, panel 1 of 4.}\label{tab:hero-250-symbolic-250-1}\\
\toprule
Dataset / statistic & \shortstack{RuleFit} & \shortstack{Boosted\\stumps} & \shortstack{SLIPPER} \\
\midrule
\endfirsthead
\caption[]{Discrete baselines at $n=250$ (continued).}\\
\toprule
Dataset / statistic & \shortstack{RuleFit} & \shortstack{Boosted\\stumps} & \shortstack{SLIPPER} \\
\midrule
\endhead
\bottomrule
\endfoot
Adult & $18.1\pm 0.4$ & $\boldsymbol{17.1\pm 0.4}$ & $19.3\pm 0.7$ \\
Bank marketing & $12.1\pm 0.1$ & $\boldsymbol{11.4\pm 0.2}$ & $12.0\pm 0.3$ \\
College & $\boldsymbol{15.6\pm 0.6}$ & $16.1\pm 0.6$ & $17.2\pm 1.2$ \\
COMPAS & $33.7\pm 0.2$ & $\boldsymbol{33.1\pm 0.4}$ & $34.3\pm 0.5$ \\
HELOC & $31.0\pm 0.4$ & $\boldsymbol{30.6\pm 0.3}$ & $32.7\pm 0.7$ \\
Stroke & $6.4\pm 0.3$ & $6.4\pm 0.4$ & $\boldsymbol{5.9\pm 0.4}$ \\
Telco churn & $\boldsymbol{21.7\pm 0.4}$ & $22.6\pm 0.4$ & $24.1\pm 0.6$ \\
Water potability & $39.0\pm 0.5$ & $42.2\pm 0.7$ & $\boldsymbol{38.9\pm 0.3}$ \\
\midrule
Macro & $\boldsymbol{22.2\pm 0.1}$ & $22.4\pm 0.2$ & $23.1\pm 0.2$ \\
Mean rank & $3.1$ & $\boldsymbol{2.9}$ & $4.1$ \\
\end{longtable}

\begin{longtable}{@{}lrrr@{}}
\caption{Discrete baselines at $n=250$, panel 2 of 4.}\label{tab:hero-250-symbolic-250-2}\\
\toprule
Dataset / statistic & \shortstack{CORELS} & \shortstack{FIGS} & \shortstack{TAO tree} \\
\midrule
\endfirsthead
\caption[]{Discrete baselines at $n=250$ (continued).}\\
\toprule
Dataset / statistic & \shortstack{CORELS} & \shortstack{FIGS} & \shortstack{TAO tree} \\
\midrule
\endhead
\bottomrule
\endfoot
Adult & $\boldsymbol{19.5\pm 0.7}$ & $20.3\pm 0.5$ & $20.2\pm 0.6$ \\
Bank marketing & $\boldsymbol{11.3\pm 0.2}$ & $13.5\pm 0.3$ & $13.7\pm 0.3$ \\
College & $24.9\pm 0.4$ & $\boldsymbol{16.4\pm 0.5}$ & $19.1\pm 0.7$ \\
COMPAS & $36.0\pm 0.7$ & $\boldsymbol{35.6\pm 0.6}$ & $36.0\pm 0.6$ \\
HELOC & $\boldsymbol{33.1\pm 0.5}$ & $34.3\pm 0.6$ & $34.5\pm 0.5$ \\
Stroke & $\boldsymbol{4.9\pm 0.0}$ & $8.3\pm 0.3$ & $8.9\pm 0.3$ \\
Telco churn & $\boldsymbol{22.8\pm 0.5}$ & $24.0\pm 0.5$ & $23.7\pm 0.6$ \\
Water potability & $\boldsymbol{42.0\pm 0.6}$ & $42.4\pm 0.9$ & $44.1\pm 1.3$ \\
\midrule
Macro & $\boldsymbol{24.3\pm 0.2}$ & $24.4\pm 0.2$ & $25.0\pm 0.2$ \\
Mean rank & $\boldsymbol{4.4}$ & $6.3$ & $6.9$ \\
\end{longtable}

\begin{longtable}{@{}lrrr@{}}
\caption{Discrete baselines at $n=250$, panel 3 of 4.}\label{tab:hero-250-symbolic-250-3}\\
\toprule
Dataset / statistic & \shortstack{Greedy\\rule list} & \shortstack{HS tree} & \shortstack{FPSkope} \\
\midrule
\endfirsthead
\caption[]{Discrete baselines at $n=250$ (continued).}\\
\toprule
Dataset / statistic & \shortstack{Greedy\\rule list} & \shortstack{HS tree} & \shortstack{FPSkope} \\
\midrule
\endhead
\bottomrule
\endfoot
Adult & $25.4\pm 1.0$ & $20.6\pm 0.5$ & $\boldsymbol{20.5\pm 0.6}$ \\
Bank marketing & $\boldsymbol{11.5\pm 0.2}$ & $14.4\pm 0.3$ & $11.7\pm 0.0$ \\
College & $23.4\pm 0.8$ & $\boldsymbol{19.0\pm 0.5}$ & $50.0\pm 0.0$ \\
COMPAS & $\boldsymbol{36.2\pm 0.8}$ & $37.6\pm 0.5$ & $41.3\pm 0.6$ \\
HELOC & $\boldsymbol{31.5\pm 0.5}$ & $35.6\pm 0.6$ & $47.6\pm 0.1$ \\
Stroke & $5.6\pm 0.6$ & $9.3\pm 0.4$ & $\boldsymbol{4.9\pm 0.0}$ \\
Telco churn & $26.9\pm 0.6$ & $23.8\pm 0.6$ & $\boldsymbol{22.6\pm 0.2}$ \\
Water potability & $40.6\pm 1.0$ & $44.7\pm 1.0$ & $\boldsymbol{40.0\pm 0.6}$ \\
\midrule
Macro & $\boldsymbol{25.1\pm 0.3}$ & $25.6\pm 0.2$ & $29.8\pm 0.1$ \\
Mean rank & $6.0$ & $8.1$ & $\boldsymbol{5.9}$ \\
\end{longtable}

\begin{longtable}{@{}lr@{}}
\caption{Discrete baselines at $n=250$, panel 4 of 4.}\label{tab:hero-250-symbolic-250-4}\\
\toprule
Dataset / statistic & \shortstack{SkopeRules} \\
\midrule
\endfirsthead
\caption[]{Discrete baselines at $n=250$ (continued).}\\
\toprule
Dataset / statistic & \shortstack{SkopeRules} \\
\midrule
\endhead
\bottomrule
\endfoot
Adult & $22.8\pm 0.6$ \\
Bank marketing & $11.7\pm 0.2$ \\
College & $50.0\pm 0.0$ \\
COMPAS & $44.9\pm 0.1$ \\
HELOC & $47.8\pm 0.0$ \\
Stroke & $6.1\pm 0.5$ \\
Telco churn & $25.1\pm 0.7$ \\
Water potability & $39.0\pm 0.1$ \\
\midrule
Macro & $30.9\pm 0.1$ \\
Mean rank & $7.3$ \\
\end{longtable}

% Imported from source/tables/hero-1200-symbolic-corrected.md
\begin{longtable}{@{}lrrr@{}}
\caption{Discrete baselines at $n=1200$, panel 1 of 4.}\label{tab:hero-1200-symbolic-corrected-1200-1}\\
\toprule
Dataset / statistic & \shortstack{Boosted\\stumps} & \shortstack{RuleFit} & \shortstack{FIGS} \\
\midrule
\endfirsthead
\caption[]{Discrete baselines at $n=1200$ (continued).}\\
\toprule
Dataset / statistic & \shortstack{Boosted\\stumps} & \shortstack{RuleFit} & \shortstack{FIGS} \\
\midrule
\endhead
\bottomrule
\endfoot
Adult & $15.8\pm 0.1$ & $\boldsymbol{15.6\pm 0.2}$ & $16.1\pm 0.2$ \\
Bank marketing & $\boldsymbol{10.7\pm 0.0}$ & $11.3\pm 0.1$ & $11.3\pm 0.1$ \\
COMPAS & $32.9\pm 0.4$ & $32.8\pm 0.3$ & $\boldsymbol{32.4\pm 0.3}$ \\
HELOC & $28.9\pm 0.3$ & $\boldsymbol{28.4\pm 0.1}$ & $29.5\pm 0.3$ \\
Stroke & $\boldsymbol{4.9\pm 0.0}$ & $\boldsymbol{4.9\pm 0.0}$ & $5.8\pm 0.2$  \\
Telco churn & $\boldsymbol{20.9\pm 0.2}$ & $21.5\pm 0.1$ & $21.5\pm 0.3$ \\
Water potability & $37.9\pm 0.3$ & $38.1\pm 0.2$ & $\boldsymbol{37.4\pm 0.6}$ \\
\midrule
Macro & $\boldsymbol{21.7\pm 0.1}$ & $21.8\pm 0.1$ & $22.0\pm 0.1$ \\
Mean rank & $\boldsymbol{2.6}$ & $3.4$ & $3.7$ \\
\end{longtable}

\begin{longtable}{@{}lrrr@{}}
\caption{Discrete baselines at $n=1200$, panel 2 of 4.}\label{tab:hero-1200-symbolic-corrected-1200-2}\\
\toprule
Dataset / statistic & \shortstack{SLIPPER} & \shortstack{TAO tree} & \shortstack{HS tree} \\
\midrule
\endfirsthead
\caption[]{Discrete baselines at $n=1200$ (continued).}\\
\toprule
Dataset / statistic & \shortstack{SLIPPER} & \shortstack{TAO tree} & \shortstack{HS tree} \\
\midrule
\endhead
\bottomrule
\endfoot
Adult & $\boldsymbol{15.6\pm 0.1}$ & $16.1\pm 0.2$ & $16.4\pm 0.2$ \\
Bank marketing & $\boldsymbol{11.1\pm 0.1}$ & $11.4\pm 0.1$ & $11.7\pm 0.2$ \\
COMPAS & $\boldsymbol{32.9\pm 0.4}$ & $33.0\pm 0.4$ & $33.3\pm 0.4$ \\
HELOC & $\boldsymbol{30.5\pm 0.7}$ & $30.6\pm 0.4$ & $31.4\pm 0.4$ \\
Stroke & $\boldsymbol{4.9\pm 0.0}$ & $5.8\pm 0.2$ & $6.3\pm 0.3$ \\
Telco churn & $\boldsymbol{21.8\pm 0.2}$ & $22.0\pm 0.3$ & $21.9\pm 0.4$ \\
Water potability & $\boldsymbol{37.5\pm 0.3}$ & $37.7\pm 0.6$ & $38.6\pm 0.4$ \\
\midrule
Macro & $\boldsymbol{22.0\pm 0.1}$ & $22.3\pm 0.1$ & $22.8\pm 0.1$ \\
Mean rank & $\boldsymbol{3.1}$ & $5.9$ & $7.4$ \\
\end{longtable}

\begin{longtable}{@{}lrrr@{}}
\caption{Discrete baselines at $n=1200$, panel 3 of 4.}\label{tab:hero-1200-symbolic-corrected-1200-3}\\
\toprule
Dataset / statistic & \shortstack{CORELS} & \shortstack{Greedy\\rule list} & \shortstack{FPSkope} \\
\midrule
\endfirsthead
\caption[]{Discrete baselines at $n=1200$ (continued).}\\
\toprule
Dataset / statistic & \shortstack{CORELS} & \shortstack{Greedy\\rule list} & \shortstack{FPSkope} \\
\midrule
\endhead
\bottomrule
\endfoot
Adult & $\boldsymbol{17.8\pm 0.0}$ & $22.5\pm 0.5$ & $19.2\pm 0.3$ \\
Bank marketing & $11.1\pm 0.1$ & $\boldsymbol{11.0\pm 0.1}$ & $11.7\pm 0.0$ \\
COMPAS & $\boldsymbol{34.8\pm 0.3}$ & $36.1\pm 0.5$ & $39.8\pm 0.6$ \\
HELOC & $31.6\pm 0.3$ & $\boldsymbol{30.3\pm 0.2}$ & $47.8\pm 0.0$ \\
Stroke & $\boldsymbol{4.9\pm 0.0}$ & $\boldsymbol{4.9\pm 0.0}$ & $\boldsymbol{4.9\pm 0.0}$ \\
Telco churn & $22.5\pm 0.5$ & $26.5\pm 0.0$ & $\boldsymbol{21.7\pm 0.1}$ \\
Water potability & $38.3\pm 0.1$ & $\boldsymbol{37.8\pm 0.4}$ & $39.0\pm 0.0$ \\
\midrule
Macro & $\boldsymbol{23.0\pm 0.1}$ & $24.2\pm 0.1$ & $26.3\pm 0.1$ \\
Mean rank & $6.4$ & $\boldsymbol{5.8}$ & $7.8$ \\
\end{longtable}

\begin{longtable}{@{}lr@{}}
\caption{Discrete baselines at $n=1200$, panel 4 of 4.}\label{tab:hero-1200-symbolic-corrected-1200-4}\\
\toprule
Dataset / statistic & \shortstack{SkopeRules} \\
\midrule
\endfirsthead
\caption[]{Discrete baselines at $n=1200$ (continued).}\\
\toprule
Dataset / statistic & \shortstack{SkopeRules} \\
\midrule
\endhead
\bottomrule
\endfoot
Adult & $23.8\pm 0.2$ \\
Bank marketing & $11.6\pm 0.1$ \\
COMPAS & $45.1\pm 0.0$ \\
HELOC & $47.8\pm 0.0$ \\
Stroke & $5.4\pm 0.2$ \\
Telco churn & $23.2\pm 0.5$ \\
Water potability & $38.9\pm 0.1$ \\
\midrule
Macro & $28.0\pm 0.1$ \\
Mean rank & $8.9$ \\
\end{longtable}

\subsection{CLIPPER and SLIPPER}
\label{app:slopslip}
% Imported from source/tables/hero-250-slopslip.md
\begin{longtable}{@{}lrrr@{}}
\caption{CLIPPER and SLIPPER at $n=250$, panel 1 of 2.}\label{tab:hero-250-slopslip-250-1}\\
\toprule
Dataset / statistic & \shortstack{CLIPPER\\suggestions} & \shortstack{CLIPPER\\no tools} & \shortstack{CLIPPER\\tools} \\
\midrule
\endfirsthead
\caption[]{CLIPPER and SLIPPER at $n=250$ (continued).}\\
\toprule
Dataset / statistic & \shortstack{CLIPPER\\suggestions} & \shortstack{CLIPPER\\no tools} & \shortstack{CLIPPER\\tools} \\
\midrule
\endhead
\bottomrule
\endfoot
Adult & $\boldsymbol{17.4\pm 0.5}$ & $17.8\pm 0.9$ & $17.6\pm 0.7$ \\
Bank marketing & $11.3\pm 0.2$ & $\boldsymbol{11.0\pm 0.1}$ & $11.3\pm 0.2$ \\
College & $\boldsymbol{17.2\pm 0.6}$ & $18.8\pm 0.9$ & $19.0\pm 0.4$ \\
COMPAS & $\boldsymbol{32.5\pm 0.4}$ & $33.5\pm 0.3$ & $33.5\pm 0.3$ \\
HELOC & $31.5\pm 0.3$ & $\boldsymbol{31.2\pm 0.3}$ & $33.1\pm 0.4$ \\
Stroke & $6.2\pm 0.5$ & $\boldsymbol{5.4\pm 0.3}$ & $6.1\pm 0.4$ \\
Telco churn & $22.4\pm 0.6$ & $23.4\pm 0.2$ & $\boldsymbol{22.2\pm 0.4}$ \\
Water potability & $41.3\pm 1.0$ & $39.9\pm 0.4$ & $\boldsymbol{38.8\pm 0.7}$ \\
\midrule
Macro & $\boldsymbol{22.5\pm 0.2}$ & $22.6\pm 0.2$ & $22.7\pm 0.2$ \\
Mean rank & $2.9$ & $\boldsymbol{2.5}$ & $3.0$ \\
\end{longtable}

\begin{longtable}{@{}lrrr@{}}
\caption{CLIPPER and SLIPPER at $n=250$, panel 2 of 2.}\label{tab:hero-250-slopslip-250-2}\\
\toprule
Dataset / statistic & \shortstack{SLIPPER\\$M=100$} & \shortstack{CLIPPER\\suggestions\\+ tools} & \shortstack{SLIPPER} \\
\midrule
\endfirsthead
\caption[]{CLIPPER and SLIPPER at $n=250$ (continued).}\\
\toprule
Dataset / statistic & \shortstack{SLIPPER\\$M=100$} & \shortstack{CLIPPER\\suggestions\\+ tools} & \shortstack{SLIPPER} \\
\midrule
\endhead
\bottomrule
\endfoot
Adult & $20.2\pm 1.0$ & $\boldsymbol{17.7\pm 0.5}$ & $19.3\pm 0.7$ \\
Bank marketing & $\boldsymbol{11.6\pm 0.0}$ & $11.8\pm 0.3$ & $12.0\pm 0.3$ \\
College & $\boldsymbol{16.3\pm 0.5}$ & $16.6\pm 0.6$ & $17.2\pm 1.2$ \\
COMPAS & $34.6\pm 0.6$ & $\boldsymbol{34.1\pm 0.4}$ & $34.3\pm 0.5$ \\
HELOC & $\boldsymbol{32.4\pm 0.9}$ & $33.5\pm 0.3$ & $32.7\pm 0.7$ \\
Stroke & $\boldsymbol{5.6\pm 0.4}$ & $6.5\pm 0.6$ & $5.9\pm 0.4$ \\
Telco churn & $\boldsymbol{23.4\pm 0.4}$ & $23.7\pm 0.9$ & $24.1\pm 0.6$ \\
Water potability & $39.9\pm 0.4$ & $40.4\pm 0.6$ & $\boldsymbol{38.9\pm 0.3}$ \\
\midrule
Macro & $\boldsymbol{23.0\pm 0.2}$ & $\boldsymbol{23.0\pm 0.2}$ & $23.1\pm 0.2$  \\
Mean rank & $\boldsymbol{3.8}$ & $4.5$ & $4.4$ \\
\end{longtable}

% Imported from source/tables/hero-1200-slopslip.md
\begin{longtable}{@{}lrrr@{}}
\caption{CLIPPER and SLIPPER at $n=1200$, panel 1 of 2.}\label{tab:hero-1200-slopslip-1200-1}\\
\toprule
Dataset / statistic & \shortstack{CLIPPER\\suggestions} & \shortstack{CLIPPER\\suggestions\\+ tools} & \shortstack{SLIPPER} \\
\midrule
\endfirsthead
\caption[]{CLIPPER and SLIPPER at $n=1200$ (continued).}\\
\toprule
Dataset / statistic & \shortstack{CLIPPER\\suggestions} & \shortstack{CLIPPER\\suggestions\\+ tools} & \shortstack{SLIPPER} \\
\midrule
\endhead
\bottomrule
\endfoot
Adult & $\boldsymbol{15.4\pm 0.1}$ & $\boldsymbol{15.4\pm 0.1}$ & $15.6\pm 0.1$  \\
Bank marketing & $\boldsymbol{10.9\pm 0.1}$ & $11.0\pm 0.1$ & $11.1\pm 0.1$ \\
COMPAS & $\boldsymbol{31.8\pm 0.2}$ & $32.0\pm 0.2$ & $32.9\pm 0.4$ \\
HELOC & $\boldsymbol{29.6\pm 0.3}$ & $30.3\pm 0.4$ & $30.5\pm 0.7$ \\
Stroke & $5.1\pm 0.2$ & $5.3\pm 0.2$ & $\boldsymbol{4.9\pm 0.0}$ \\
Telco churn & $\boldsymbol{21.2\pm 0.3}$ & $21.4\pm 0.3$ & $21.8\pm 0.2$ \\
Water potability & $38.1\pm 0.3$ & $37.8\pm 0.4$ & $\boldsymbol{37.5\pm 0.3}$ \\
\midrule
Macro & $\boldsymbol{21.7\pm 0.1}$ & $21.9\pm 0.1$ & $22.0\pm 0.1$ \\
Mean rank & $\boldsymbol{2.0}$ & $2.9$ & $3.1$ \\
\end{longtable}

\begin{longtable}{@{}lrrr@{}}
\caption{CLIPPER and SLIPPER at $n=1200$, panel 2 of 2.}\label{tab:hero-1200-slopslip-1200-2}\\
\toprule
Dataset / statistic & \shortstack{CLIPPER\\tools} & \shortstack{CLIPPER\\no tools} & \shortstack{SLIPPER\\$M=100$} \\
\midrule
\endfirsthead
\caption[]{CLIPPER and SLIPPER at $n=1200$ (continued).}\\
\toprule
Dataset / statistic & \shortstack{CLIPPER\\tools} & \shortstack{CLIPPER\\no tools} & \shortstack{SLIPPER\\$M=100$} \\
\midrule
\endhead
\bottomrule
\endfoot
Adult & $\boldsymbol{16.0\pm 0.2}$ & $16.1\pm 0.1$ & $17.9\pm 0.7$ \\
Bank marketing & $11.0\pm 0.1$ & $\boldsymbol{10.8\pm 0.0}$ & $11.7\pm 0.0$ \\
COMPAS & $\boldsymbol{32.8\pm 0.2}$ & $\boldsymbol{32.8\pm 0.1}$ & $33.2\pm 0.4$  \\
HELOC & $32.4\pm 0.5$ & $\boldsymbol{30.6\pm 0.3}$ & $33.8\pm 0.6$ \\
Stroke & $5.1\pm 0.2$ & $5.1\pm 0.2$ & $\boldsymbol{4.9\pm 0.0}$ \\
Telco churn & $\boldsymbol{22.1\pm 0.4}$ & $23.5\pm 0.1$ & $23.6\pm 0.6$ \\
Water potability & $\boldsymbol{37.4\pm 0.4}$ & $38.8\pm 0.2$ & $38.7\pm 0.2$ \\
\midrule
Macro & $\boldsymbol{22.4\pm 0.1}$ & $22.5\pm 0.1$ & $23.4\pm 0.2$ \\
Mean rank & $\boldsymbol{3.9}$ & $\boldsymbol{3.9}$ & $5.3$ \\
\end{longtable}

\subsection{Alternative harnesses}
\label{app:harness}
% Imported from source/tables/hero-250-harness.md
\begin{longtable}{@{}lrrr@{}}
\caption{Alternative harnesses at $n=250$, panel 1 of 3.}\label{tab:hero-250-harness-250-1}\\
\toprule
Dataset / statistic & \shortstack{CLIPPER\\tools} & \shortstack{CLIPPER\\set cover} & \shortstack{CLIPPER\\suggestions\\+ tools} \\
\midrule
\endfirsthead
\caption[]{Alternative harnesses at $n=250$ (continued).}\\
\toprule
Dataset / statistic & \shortstack{CLIPPER\\tools} & \shortstack{CLIPPER\\set cover} & \shortstack{CLIPPER\\suggestions\\+ tools} \\
\midrule
\endhead
\bottomrule
\endfoot
Adult & $\boldsymbol{17.6\pm 0.7}$ & $17.7\pm 0.6$ & $17.7\pm 0.5$ \\
Bank marketing & $\boldsymbol{11.3\pm 0.2}$ & $\boldsymbol{11.3\pm 0.3}$ & $11.8\pm 0.3$  \\
College & $19.0\pm 0.4$ & $22.3\pm 0.8$ & $\boldsymbol{16.6\pm 0.6}$ \\
COMPAS & $33.5\pm 0.3$ & $\boldsymbol{33.3\pm 0.2}$ & $34.1\pm 0.4$ \\
HELOC & $33.1\pm 0.4$ & $\boldsymbol{31.2\pm 0.4}$ & $33.5\pm 0.3$ \\
Stroke & $6.1\pm 0.4$ & $\boldsymbol{5.4\pm 0.4}$ & $6.5\pm 0.6$ \\
Telco churn & $\boldsymbol{22.2\pm 0.4}$ & $22.9\pm 0.3$ & $23.7\pm 0.9$ \\
Water potability & $\boldsymbol{38.8\pm 0.7}$ & $39.5\pm 0.3$ & $40.4\pm 0.6$ \\
\midrule
Macro & $\boldsymbol{22.7\pm 0.2}$ & $22.9\pm 0.2$ & $23.0\pm 0.2$ \\
Mean rank & $\boldsymbol{2.9}$ & $3.0$ & $4.8$ \\
\end{longtable}

\begin{longtable}{@{}lrrr@{}}
\caption{Alternative harnesses at $n=250$, panel 2 of 3.}\label{tab:hero-250-harness-250-2}\\
\toprule
Dataset / statistic & \shortstack{SLIPPER} & \shortstack{One-shot\\CLIPPER} & \shortstack{Discrete\\set cover} \\
\midrule
\endfirsthead
\caption[]{Alternative harnesses at $n=250$ (continued).}\\
\toprule
Dataset / statistic & \shortstack{SLIPPER} & \shortstack{One-shot\\CLIPPER} & \shortstack{Discrete\\set cover} \\
\midrule
\endhead
\bottomrule
\endfoot
Adult & $19.3\pm 0.7$ & $\boldsymbol{18.2\pm 0.9}$ & $25.4\pm 0.6$ \\
Bank marketing & $12.0\pm 0.3$ & $11.7\pm 0.2$ & $\boldsymbol{11.4\pm 0.1}$ \\
College & $\boldsymbol{17.2\pm 1.2}$ & $28.4\pm 1.2$ & $20.3\pm 0.8$ \\
COMPAS & $\boldsymbol{34.3\pm 0.5}$ & $35.3\pm 1.3$ & $36.7\pm 1.1$ \\
HELOC & $32.7\pm 0.7$ & $\boldsymbol{32.3\pm 0.4}$ & $33.3\pm 0.9$ \\
Stroke & $5.9\pm 0.4$ & $5.3\pm 0.4$ & $\boldsymbol{4.9\pm 0.0}$ \\
Telco churn & $\boldsymbol{24.1\pm 0.6}$ & $24.9\pm 0.5$ & $25.5\pm 1.0$ \\
Water potability & $\boldsymbol{38.9\pm 0.3}$ & $39.9\pm 0.4$ & $42.5\pm 1.1$ \\
\midrule
Macro & $\boldsymbol{23.1\pm 0.2}$ & $24.5\pm 0.3$ & $25.0\pm 0.3$ \\
Mean rank & $\boldsymbol{4.4}$ & $4.8$ & $5.8$ \\
\end{longtable}

\begin{longtable}{@{}lrr@{}}
\caption{Alternative harnesses at $n=250$, panel 3 of 3.}\label{tab:hero-250-harness-250-3}\\
\toprule
Dataset / statistic & \shortstack{Suggestions\\set cover} & \shortstack{One-shot\\suggestions} \\
\midrule
\endfirsthead
\caption[]{Alternative harnesses at $n=250$ (continued).}\\
\toprule
Dataset / statistic & \shortstack{Suggestions\\set cover} & \shortstack{One-shot\\suggestions} \\
\midrule
\endhead
\bottomrule
\endfoot
Adult & $\boldsymbol{20.5\pm 1.2}$ & $23.4\pm 0.7$ \\
Bank marketing & $\boldsymbol{11.1\pm 0.2}$ & $11.7\pm 0.0$ \\
College & $\boldsymbol{36.9\pm 2.5}$ & $40.0\pm 3.4$ \\
COMPAS & $\boldsymbol{36.4\pm 1.1}$ & $41.7\pm 1.4$ \\
HELOC & $\boldsymbol{30.9\pm 0.3}$ & $38.1\pm 2.3$ \\
Stroke & $\boldsymbol{4.9\pm 0.0}$ & $\boldsymbol{4.9\pm 0.0}$ \\
Telco churn & $\boldsymbol{25.2\pm 0.4}$ & $27.0\pm 0.6$ \\
Water potability & $39.6\pm 0.4$ & $\boldsymbol{39.3\pm 0.3}$ \\
\midrule
Macro & $\boldsymbol{25.7\pm 0.4}$ & $28.2\pm 0.6$ \\
Mean rank & $\boldsymbol{4.3}$ & $6.3$ \\
\end{longtable}

% Imported from source/tables/hero-1200-harness.md
\begin{longtable}{@{}lrrr@{}}
\caption{Alternative harnesses at $n=1200$, panel 1 of 3.}\label{tab:hero-1200-harness-1200-1}\\
\toprule
Dataset / statistic & \shortstack{CLIPPER\\suggestions\\+ tools} & \shortstack{SLIPPER} & \shortstack{CLIPPER\\tools} \\
\midrule
\endfirsthead
\caption[]{Alternative harnesses at $n=1200$ (continued).}\\
\toprule
Dataset / statistic & \shortstack{CLIPPER\\suggestions\\+ tools} & \shortstack{SLIPPER} & \shortstack{CLIPPER\\tools} \\
\midrule
\endhead
\bottomrule
\endfoot
Adult & $\boldsymbol{15.4\pm 0.1}$ & $15.6\pm 0.1$ & $16.0\pm 0.2$ \\
Bank marketing & $\boldsymbol{11.0\pm 0.1}$ & $11.1\pm 0.1$ & $\boldsymbol{11.0\pm 0.1}$  \\
COMPAS & $\boldsymbol{32.0\pm 0.2}$ & $32.9\pm 0.4$ & $32.8\pm 0.2$ \\
HELOC & $\boldsymbol{30.3\pm 0.4}$ & $30.5\pm 0.7$ & $32.4\pm 0.5$ \\
Stroke & $5.3\pm 0.2$ & $\boldsymbol{4.9\pm 0.0}$ & $5.1\pm 0.2$ \\
Telco churn & $\boldsymbol{21.4\pm 0.3}$ & $21.8\pm 0.2$ & $22.1\pm 0.4$ \\
Water potability & $37.8\pm 0.4$ & $37.5\pm 0.3$ & $\boldsymbol{37.4\pm 0.4}$ \\
\midrule
Macro & $\boldsymbol{21.9\pm 0.1}$ & $22.0\pm 0.1$ & $22.4\pm 0.1$ \\
Mean rank & $\boldsymbol{2.3}$ & $2.4$ & $3.4$ \\
\end{longtable}

\begin{longtable}{@{}lrrr@{}}
\caption{Alternative harnesses at $n=1200$, panel 2 of 3.}\label{tab:hero-1200-harness-1200-2}\\
\toprule
Dataset / statistic & \shortstack{CLIPPER\\set cover} & \shortstack{Discrete\\set cover} & \shortstack{Suggestions\\set cover} \\
\midrule
\endfirsthead
\caption[]{Alternative harnesses at $n=1200$ (continued).}\\
\toprule
Dataset / statistic & \shortstack{CLIPPER\\set cover} & \shortstack{Discrete\\set cover} & \shortstack{Suggestions\\set cover} \\
\midrule
\endhead
\bottomrule
\endfoot
Adult & $\boldsymbol{16.3\pm 0.2}$ & $21.4\pm 0.6$ & $21.1\pm 1.0$ \\
Bank marketing & $\boldsymbol{11.0\pm 0.1}$ & $11.1\pm 0.1$ & $11.4\pm 0.1$ \\
COMPAS & $\boldsymbol{33.2\pm 0.4}$ & $34.4\pm 0.6$ & $36.5\pm 1.2$ \\
HELOC & $\boldsymbol{30.5\pm 0.2}$ & $30.8\pm 0.4$ & $\boldsymbol{30.5\pm 0.2}$  \\
Stroke & $\boldsymbol{4.9\pm 0.0}$ & $\boldsymbol{4.9\pm 0.0}$ & $\boldsymbol{4.9\pm 0.0}$ \\
Telco churn & $\boldsymbol{22.3\pm 0.2}$ & $26.0\pm 0.5$ & $24.7\pm 0.3$ \\
Water potability & $39.1\pm 0.2$ & $39.0\pm 0.6$ & $\boldsymbol{38.9\pm 0.1}$ \\
\midrule
Macro & $\boldsymbol{22.5\pm 0.1}$ & $23.9\pm 0.2$ & $24.0\pm 0.2$ \\
Mean rank & $\boldsymbol{4.4}$ & $5.4$ & $4.9$ \\
\end{longtable}

\begin{longtable}{@{}lrr@{}}
\caption{Alternative harnesses at $n=1200$, panel 3 of 3.}\label{tab:hero-1200-harness-1200-3}\\
\toprule
Dataset / statistic & \shortstack{One-shot\\CLIPPER} & \shortstack{One-shot\\suggestions} \\
\midrule
\endfirsthead
\caption[]{Alternative harnesses at $n=1200$ (continued).}\\
\toprule
Dataset / statistic & \shortstack{One-shot\\CLIPPER} & \shortstack{One-shot\\suggestions} \\
\midrule
\endhead
\bottomrule
\endfoot
Adult & $\boldsymbol{18.8\pm 1.0}$ & $24.1\pm 0.0$ \\
Bank marketing & $\boldsymbol{11.3\pm 0.1}$ & $11.7\pm 0.0$ \\
COMPAS & $\boldsymbol{36.5\pm 1.2}$ & $43.9\pm 0.8$ \\
HELOC & $37.7\pm 2.4$ & $\boldsymbol{37.5\pm 2.2}$ \\
Stroke & $\boldsymbol{4.9\pm 0.0}$ & $\boldsymbol{4.9\pm 0.0}$ \\
Telco churn & $\boldsymbol{25.6\pm 0.5}$ & $26.7\pm 0.2$ \\
Water potability & $\boldsymbol{39.0\pm 0.0}$ & $\boldsymbol{39.0\pm 0.0}$ \\
\midrule
Macro & $\boldsymbol{24.8\pm 0.4}$ & $26.8\pm 0.3$ \\
Mean rank & $\boldsymbol{6.1}$ & $7.1$ \\
\end{longtable}

\subsection{Complete summary learning-curve results}
\label{app:punchline}
% Imported from source/tables/hero-punchline-curve.md
\begin{longtable}{@{}lrrr@{}}
\caption{Complete summary learning-curve results at $n=100$, panel 1 of 2.}\label{tab:hero-punchline-curve-100-1}\\
\toprule
Dataset / statistic & \shortstack{CLIPPER\\suggestions} & \shortstack{CLIPPER\\tools} & \shortstack{CLIPPER\\no tools} \\
\midrule
\endfirsthead
\caption[]{Complete summary learning-curve results at $n=100$ (continued).}\\
\toprule
Dataset / statistic & \shortstack{CLIPPER\\suggestions} & \shortstack{CLIPPER\\tools} & \shortstack{CLIPPER\\no tools} \\
\midrule
\endhead
\bottomrule
\endfoot
Adult & $19.0\pm 0.5$ & $\boldsymbol{18.1\pm 0.5}$ & $18.3\pm 0.7$ \\
Bank marketing & $12.1\pm 0.3$ & $11.8\pm 0.3$ & $\boldsymbol{11.5\pm 0.2}$ \\
College & $\boldsymbol{20.4\pm 1.3}$ & $22.9\pm 1.0$ & $25.7\pm 1.4$ \\
COMPAS & $\boldsymbol{36.4\pm 1.0}$ & $37.6\pm 1.5$ & $37.6\pm 1.1$ \\
HELOC & $36.9\pm 1.0$ & $36.4\pm 1.3$ & $\boldsymbol{35.9\pm 1.9}$ \\
Stroke & $5.2\pm 0.2$ & $5.0\pm 0.1$ & $\boldsymbol{4.9\pm 0.0}$ \\
Telco churn & $25.2\pm 1.1$ & $\boldsymbol{23.6\pm 0.5}$ & $24.7\pm 0.5$ \\
Water potability & $41.2\pm 0.7$ & $41.3\pm 0.7$ & $\boldsymbol{40.5\pm 0.6}$ \\
\midrule
Macro & $\boldsymbol{24.5\pm 0.3}$ & $24.6\pm 0.3$ & $24.9\pm 0.4$ \\
Mean rank & $3.0$ & $2.4$ & $\boldsymbol{2.1}$ \\
\end{longtable}

\begin{longtable}{@{}lrr@{}}
\caption{Complete summary learning-curve results at $n=100$, panel 2 of 2.}\label{tab:hero-punchline-curve-100-2}\\
\toprule
Dataset / statistic & \shortstack{CLIPPER\\suggestions\\+ tools} & \shortstack{SLIPPER} \\
\midrule
\endfirsthead
\caption[]{Complete summary learning-curve results at $n=100$ (continued).}\\
\toprule
Dataset / statistic & \shortstack{CLIPPER\\suggestions\\+ tools} & \shortstack{SLIPPER} \\
\midrule
\endhead
\bottomrule
\endfoot
Adult & $\boldsymbol{19.6\pm 0.4}$ & $20.9\pm 0.7$ \\
Bank marketing & $\boldsymbol{13.3\pm 0.6}$ & $13.8\pm 0.6$ \\
College & $\boldsymbol{20.2\pm 1.0}$ & $20.3\pm 1.1$ \\
COMPAS & $\boldsymbol{37.7\pm 1.0}$ & $39.1\pm 0.9$ \\
HELOC & $\boldsymbol{37.1\pm 1.0}$ & $39.0\pm 1.3$ \\
Stroke & $6.5\pm 0.4$ & $\boldsymbol{5.3\pm 0.2}$ \\
Telco churn & $\boldsymbol{23.9\pm 0.5}$ & $24.5\pm 0.7$ \\
Water potability & $42.0\pm 0.9$ & $\boldsymbol{41.0\pm 0.8}$ \\
\midrule
Macro & $\boldsymbol{25.0\pm 0.3}$ & $25.5\pm 0.3$ \\
Mean rank & $\boldsymbol{3.6}$ & $3.9$ \\
\end{longtable}

\begin{longtable}{@{}lrrr@{}}
\caption{Complete summary learning-curve results at $n=175$, panel 1 of 2.}\label{tab:hero-punchline-curve-175-1}\\
\toprule
Dataset / statistic & \shortstack{CLIPPER\\suggestions} & \shortstack{SLIPPER} & \shortstack{CLIPPER\\no tools} \\
\midrule
\endfirsthead
\caption[]{Complete summary learning-curve results at $n=175$ (continued).}\\
\toprule
Dataset / statistic & \shortstack{CLIPPER\\suggestions} & \shortstack{SLIPPER} & \shortstack{CLIPPER\\no tools} \\
\midrule
\endhead
\bottomrule
\endfoot
Adult & $18.2\pm 0.7$ & $18.9\pm 0.4$ & $\boldsymbol{18.0\pm 0.8}$ \\
Bank marketing & $12.0\pm 0.3$ & $12.5\pm 0.5$ & $\boldsymbol{11.2\pm 0.2}$ \\
College & $\boldsymbol{18.6\pm 1.0}$ & $18.8\pm 1.1$ & $22.0\pm 1.7$ \\
COMPAS & $\boldsymbol{32.8\pm 0.5}$ & $34.8\pm 0.9$ & $33.6\pm 0.4$ \\
HELOC & $\boldsymbol{32.7\pm 0.7}$ & $33.6\pm 0.8$ & $32.8\pm 1.3$ \\
Stroke & $5.2\pm 0.3$ & $\boldsymbol{5.1\pm 0.2}$ & $\boldsymbol{5.1\pm 0.2}$  \\
Telco churn & $\boldsymbol{23.5\pm 0.7}$ & $23.9\pm 0.5$ & $23.7\pm 0.3$ \\
Water potability & $40.5\pm 1.0$ & $\boldsymbol{39.3\pm 0.3}$ & $41.0\pm 0.9$ \\
\midrule
Macro & $\boldsymbol{22.9\pm 0.2}$ & $23.3\pm 0.2$ & $23.4\pm 0.3$ \\
Mean rank & $\boldsymbol{2.1}$ & $3.4$ & $2.9$ \\
\end{longtable}

\begin{longtable}{@{}lrr@{}}
\caption{Complete summary learning-curve results at $n=175$, panel 2 of 2.}\label{tab:hero-punchline-curve-175-2}\\
\toprule
Dataset / statistic & \shortstack{CLIPPER\\tools} & \shortstack{CLIPPER\\suggestions\\+ tools} \\
\midrule
\endfirsthead
\caption[]{Complete summary learning-curve results at $n=175$ (continued).}\\
\toprule
Dataset / statistic & \shortstack{CLIPPER\\tools} & \shortstack{CLIPPER\\suggestions\\+ tools} \\
\midrule
\endhead
\bottomrule
\endfoot
Adult & $\boldsymbol{16.8\pm 0.4}$ & $18.5\pm 0.7$ \\
Bank marketing & $\boldsymbol{11.6\pm 0.3}$ & $12.5\pm 0.4$ \\
College & $23.0\pm 1.6$ & $\boldsymbol{18.1\pm 0.9}$ \\
COMPAS & $\boldsymbol{33.4\pm 0.3}$ & $34.1\pm 0.6$ \\
HELOC & $34.4\pm 1.3$ & $\boldsymbol{34.2\pm 0.7}$ \\
Stroke & $\boldsymbol{5.4\pm 0.3}$ & $6.2\pm 0.4$ \\
Telco churn & $\boldsymbol{22.7\pm 0.4}$ & $23.5\pm 0.4$ \\
Water potability & $\boldsymbol{40.9\pm 0.6}$ & $\boldsymbol{40.9\pm 1.0}$  \\
\midrule
Macro & $\boldsymbol{23.5\pm 0.3}$ & $\boldsymbol{23.5\pm 0.2}$  \\
Mean rank & $\boldsymbol{2.9}$ & $3.8$ \\
\end{longtable}

\begin{longtable}{@{}lrrr@{}}
\caption{Complete summary learning-curve results at $n=250$, panel 1 of 2.}\label{tab:hero-punchline-curve-250-1}\\
\toprule
Dataset / statistic & \shortstack{CLIPPER\\suggestions} & \shortstack{CLIPPER\\no tools} & \shortstack{CLIPPER\\tools} \\
\midrule
\endfirsthead
\caption[]{Complete summary learning-curve results at $n=250$ (continued).}\\
\toprule
Dataset / statistic & \shortstack{CLIPPER\\suggestions} & \shortstack{CLIPPER\\no tools} & \shortstack{CLIPPER\\tools} \\
\midrule
\endhead
\bottomrule
\endfoot
Adult & $\boldsymbol{17.4\pm 0.5}$ & $17.8\pm 0.9$ & $17.6\pm 0.7$ \\
Bank marketing & $11.3\pm 0.2$ & $\boldsymbol{11.0\pm 0.1}$ & $11.3\pm 0.2$ \\
College & $\boldsymbol{17.2\pm 0.6}$ & $18.8\pm 0.9$ & $19.0\pm 0.4$ \\
COMPAS & $\boldsymbol{32.5\pm 0.4}$ & $33.5\pm 0.3$ & $33.5\pm 0.3$ \\
HELOC & $31.5\pm 0.3$ & $\boldsymbol{31.2\pm 0.3}$ & $33.1\pm 0.4$ \\
Stroke & $6.2\pm 0.5$ & $\boldsymbol{5.4\pm 0.3}$ & $6.1\pm 0.4$ \\
Telco churn & $22.4\pm 0.6$ & $23.4\pm 0.2$ & $\boldsymbol{22.2\pm 0.4}$ \\
Water potability & $41.3\pm 1.0$ & $39.9\pm 0.4$ & $\boldsymbol{38.8\pm 0.7}$ \\
\midrule
Macro & $\boldsymbol{22.5\pm 0.2}$ & $22.6\pm 0.2$ & $22.7\pm 0.2$ \\
Mean rank & $2.5$ & $\boldsymbol{2.4}$ & $2.6$ \\
\end{longtable}

\begin{longtable}{@{}lrr@{}}
\caption{Complete summary learning-curve results at $n=250$, panel 2 of 2.}\label{tab:hero-punchline-curve-250-2}\\
\toprule
Dataset / statistic & \shortstack{CLIPPER\\suggestions\\+ tools} & \shortstack{SLIPPER} \\
\midrule
\endfirsthead
\caption[]{Complete summary learning-curve results at $n=250$ (continued).}\\
\toprule
Dataset / statistic & \shortstack{CLIPPER\\suggestions\\+ tools} & \shortstack{SLIPPER} \\
\midrule
\endhead
\bottomrule
\endfoot
Adult & $\boldsymbol{17.7\pm 0.5}$ & $19.3\pm 0.7$ \\
Bank marketing & $\boldsymbol{11.8\pm 0.3}$ & $12.0\pm 0.3$ \\
College & $\boldsymbol{16.6\pm 0.6}$ & $17.2\pm 1.2$ \\
COMPAS & $\boldsymbol{34.1\pm 0.4}$ & $34.3\pm 0.5$ \\
HELOC & $33.5\pm 0.3$ & $\boldsymbol{32.7\pm 0.7}$ \\
Stroke & $6.5\pm 0.6$ & $\boldsymbol{5.9\pm 0.4}$ \\
Telco churn & $\boldsymbol{23.7\pm 0.9}$ & $24.1\pm 0.6$ \\
Water potability & $40.4\pm 0.6$ & $\boldsymbol{38.9\pm 0.3}$ \\
\midrule
Macro & $\boldsymbol{23.0\pm 0.2}$ & $23.1\pm 0.2$ \\
Mean rank & $\boldsymbol{3.8}$ & $\boldsymbol{3.8}$ \\
\end{longtable}

\begin{longtable}{@{}lrrr@{}}
\caption{Complete summary learning-curve results at $n=400$, panel 1 of 2.}\label{tab:hero-punchline-curve-400-1}\\
\toprule
Dataset / statistic & \shortstack{CLIPPER\\suggestions\\+ tools} & \shortstack{CLIPPER\\suggestions} & \shortstack{SLIPPER} \\
\midrule
\endfirsthead
\caption[]{Complete summary learning-curve results at $n=400$ (continued).}\\
\toprule
Dataset / statistic & \shortstack{CLIPPER\\suggestions\\+ tools} & \shortstack{CLIPPER\\suggestions} & \shortstack{SLIPPER} \\
\midrule
\endhead
\bottomrule
\endfoot
Adult & $16.4\pm 0.2$ & $\boldsymbol{15.9\pm 0.1}$ & $17.0\pm 0.4$ \\
Bank marketing & $11.8\pm 0.2$ & $\boldsymbol{11.4\pm 0.2}$ & $11.9\pm 0.3$ \\
College & $\boldsymbol{14.3\pm 0.7}$ & $16.3\pm 0.4$ & $15.2\pm 0.7$ \\
COMPAS & $34.1\pm 0.4$ & $\boldsymbol{33.1\pm 0.4}$ & $33.6\pm 0.6$ \\
HELOC & $31.5\pm 0.6$ & $\boldsymbol{30.9\pm 0.4}$ & $33.3\pm 1.4$ \\
Stroke & $5.6\pm 0.3$ & $5.5\pm 0.3$ & $\boldsymbol{5.1\pm 0.2}$ \\
Telco churn & $\boldsymbol{22.1\pm 0.3}$ & $22.2\pm 0.4$ & $23.2\pm 0.4$ \\
Water potability & $39.6\pm 0.5$ & $40.2\pm 0.8$ & $\boldsymbol{39.1\pm 0.5}$ \\
\midrule
Macro & $\boldsymbol{21.9\pm 0.2}$ & $\boldsymbol{21.9\pm 0.1}$ & $22.3\pm 0.2$  \\
Mean rank & $2.9$ & $\boldsymbol{2.4}$ & $3.1$ \\
\end{longtable}

\begin{longtable}{@{}lrr@{}}
\caption{Complete summary learning-curve results at $n=400$, panel 2 of 2.}\label{tab:hero-punchline-curve-400-2}\\
\toprule
Dataset / statistic & \shortstack{CLIPPER\\no tools} & \shortstack{CLIPPER\\tools} \\
\midrule
\endfirsthead
\caption[]{Complete summary learning-curve results at $n=400$ (continued).}\\
\toprule
Dataset / statistic & \shortstack{CLIPPER\\no tools} & \shortstack{CLIPPER\\tools} \\
\midrule
\endhead
\bottomrule
\endfoot
Adult & $\boldsymbol{16.0\pm 0.2}$ & $\boldsymbol{16.0\pm 0.2}$  \\
Bank marketing & $11.4\pm 0.2$ & $\boldsymbol{11.3\pm 0.1}$ \\
College & $\boldsymbol{17.9\pm 0.5}$ & $19.5\pm 1.0$ \\
COMPAS & $34.5\pm 0.5$ & $\boldsymbol{34.1\pm 0.4}$ \\
HELOC & $\boldsymbol{30.9\pm 0.3}$ & $32.9\pm 0.6$ \\
Stroke & $5.6\pm 0.3$ & $\boldsymbol{5.4\pm 0.3}$ \\
Telco churn & $23.2\pm 0.1$ & $\boldsymbol{22.2\pm 0.3}$ \\
Water potability & $40.3\pm 0.7$ & $\boldsymbol{39.9\pm 0.5}$ \\
\midrule
Macro & $\boldsymbol{22.5\pm 0.1}$ & $22.7\pm 0.2$ \\
Mean rank & $3.6$ & $\boldsymbol{3.0}$ \\
\end{longtable}

\begin{longtable}{@{}lrrr@{}}
\caption{Complete summary learning-curve results at $n=500$, panel 1 of 2.}\label{tab:hero-punchline-curve-500-1}\\
\toprule
Dataset / statistic & \shortstack{CLIPPER\\suggestions\\+ tools} & \shortstack{CLIPPER\\suggestions} & \shortstack{SLIPPER} \\
\midrule
\endfirsthead
\caption[]{Complete summary learning-curve results at $n=500$ (continued).}\\
\toprule
Dataset / statistic & \shortstack{CLIPPER\\suggestions\\+ tools} & \shortstack{CLIPPER\\suggestions} & \shortstack{SLIPPER} \\
\midrule
\endhead
\bottomrule
\endfoot
Adult & $16.7\pm 0.3$ & $\boldsymbol{16.2\pm 0.4}$ & $16.5\pm 0.2$ \\
Bank marketing & $11.4\pm 0.1$ & $\boldsymbol{11.1\pm 0.1}$ & $11.5\pm 0.2$ \\
College & $\boldsymbol{13.6\pm 0.6}$ & $16.7\pm 0.6$ & $15.2\pm 0.8$ \\
COMPAS & $32.9\pm 0.3$ & $\boldsymbol{32.4\pm 0.3}$ & $33.2\pm 0.4$ \\
HELOC & $31.3\pm 0.5$ & $\boldsymbol{30.8\pm 0.4}$ & $31.7\pm 0.4$ \\
Stroke & $5.3\pm 0.3$ & $5.0\pm 0.1$ & $\boldsymbol{4.9\pm 0.0}$ \\
Telco churn & $21.7\pm 0.2$ & $\boldsymbol{21.4\pm 0.2}$ & $22.2\pm 0.3$ \\
Water potability & $39.6\pm 0.6$ & $39.9\pm 0.3$ & $\boldsymbol{38.8\pm 0.2}$ \\
\midrule
Macro & $\boldsymbol{21.6\pm 0.1}$ & $21.7\pm 0.1$ & $21.8\pm 0.1$ \\
Mean rank & $3.3$ & $\boldsymbol{2.3}$ & $3.2$ \\
\end{longtable}

\begin{longtable}{@{}lrr@{}}
\caption{Complete summary learning-curve results at $n=500$, panel 2 of 2.}\label{tab:hero-punchline-curve-500-2}\\
\toprule
Dataset / statistic & \shortstack{CLIPPER\\no tools} & \shortstack{CLIPPER\\tools} \\
\midrule
\endfirsthead
\caption[]{Complete summary learning-curve results at $n=500$ (continued).}\\
\toprule
Dataset / statistic & \shortstack{CLIPPER\\no tools} & \shortstack{CLIPPER\\tools} \\
\midrule
\endhead
\bottomrule
\endfoot
Adult & $16.3\pm 0.3$ & $\boldsymbol{15.8\pm 0.2}$ \\
Bank marketing & $11.4\pm 0.2$ & $\boldsymbol{11.2\pm 0.2}$ \\
College & $\boldsymbol{18.7\pm 0.8}$ & $19.8\pm 0.5$ \\
COMPAS & $\boldsymbol{32.8\pm 0.2}$ & $33.5\pm 0.4$ \\
HELOC & $\boldsymbol{30.4\pm 0.2}$ & $32.1\pm 0.3$ \\
Stroke & $\boldsymbol{4.9\pm 0.0}$ & $5.3\pm 0.2$ \\
Telco churn & $22.9\pm 0.2$ & $\boldsymbol{21.8\pm 0.2}$ \\
Water potability & $39.6\pm 0.3$ & $\boldsymbol{38.9\pm 0.5}$ \\
\midrule
Macro & $\boldsymbol{22.1\pm 0.1}$ & $22.3\pm 0.1$ \\
Mean rank & $\boldsymbol{2.8}$ & $3.5$ \\
\end{longtable}

\begin{longtable}{@{}lrrr@{}}
\caption{Complete summary learning-curve results at $n=600$, panel 1 of 2.}\label{tab:hero-punchline-curve-600-1}\\
\toprule
Dataset / statistic & \shortstack{SLIPPER} & \shortstack{CLIPPER\\suggestions\\+ tools} & \shortstack{CLIPPER\\suggestions} \\
\midrule
\endfirsthead
\caption[]{Complete summary learning-curve results at $n=600$ (continued).}\\
\toprule
Dataset / statistic & \shortstack{SLIPPER} & \shortstack{CLIPPER\\suggestions\\+ tools} & \shortstack{CLIPPER\\suggestions} \\
\midrule
\endhead
\bottomrule
\endfoot
Adult & $16.6\pm 0.1$ & $\boldsymbol{16.0\pm 0.1}$ & $\boldsymbol{16.0\pm 0.2}$  \\
Bank marketing & $11.1\pm 0.2$ & $\boldsymbol{11.0\pm 0.1}$ & $\boldsymbol{11.0\pm 0.1}$  \\
College & $13.6\pm 0.5$ & $\boldsymbol{13.5\pm 0.5}$ & $15.7\pm 0.6$ \\
COMPAS & $32.9\pm 0.4$ & $\boldsymbol{32.3\pm 0.4}$ & $32.7\pm 0.5$ \\
HELOC & $31.2\pm 0.5$ & $31.1\pm 0.3$ & $\boldsymbol{30.6\pm 0.3}$ \\
Stroke & $\boldsymbol{4.9\pm 0.0}$ & $5.4\pm 0.3$ & $5.0\pm 0.1$ \\
Telco churn & $21.9\pm 0.2$ & $21.7\pm 0.3$ & $\boldsymbol{21.4\pm 0.2}$ \\
Water potability & $\boldsymbol{38.5\pm 0.5}$ & $39.8\pm 0.7$ & $39.7\pm 0.6$ \\
\midrule
Macro & $\boldsymbol{21.3\pm 0.1}$ & $\boldsymbol{21.3\pm 0.1}$ & $21.5\pm 0.1$ \\
Mean rank & $2.9$ & $2.6$ & $\boldsymbol{2.4}$ \\
\end{longtable}

\begin{longtable}{@{}lrr@{}}
\caption{Complete summary learning-curve results at $n=600$, panel 2 of 2.}\label{tab:hero-punchline-curve-600-2}\\
\toprule
Dataset / statistic & \shortstack{CLIPPER\\tools} & \shortstack{CLIPPER\\no tools} \\
\midrule
\endfirsthead
\caption[]{Complete summary learning-curve results at $n=600$ (continued).}\\
\toprule
Dataset / statistic & \shortstack{CLIPPER\\tools} & \shortstack{CLIPPER\\no tools} \\
\midrule
\endhead
\bottomrule
\endfoot
Adult & $\boldsymbol{16.0\pm 0.3}$ & $16.6\pm 0.4$ \\
Bank marketing & $11.0\pm 0.1$ & $\boldsymbol{10.9\pm 0.1}$ \\
College & $\boldsymbol{17.3\pm 0.5}$ & $18.0\pm 0.7$ \\
COMPAS & $\boldsymbol{33.6\pm 0.3}$ & $\boldsymbol{33.6\pm 0.3}$  \\
HELOC & $32.3\pm 0.4$ & $\boldsymbol{30.8\pm 0.4}$ \\
Stroke & $5.1\pm 0.2$ & $\boldsymbol{4.9\pm 0.0}$ \\
Telco churn & $\boldsymbol{22.0\pm 0.2}$ & $23.4\pm 0.1$ \\
Water potability & $\boldsymbol{39.4\pm 0.5}$ & $\boldsymbol{39.4\pm 0.2}$  \\
\midrule
Macro & $\boldsymbol{22.1\pm 0.1}$ & $22.2\pm 0.1$ \\
Mean rank & $3.6$ & $\boldsymbol{3.4}$ \\
\end{longtable}

\begin{longtable}{@{}lrrr@{}}
\caption{Complete summary learning-curve results at $n=800$, panel 1 of 2.}\label{tab:hero-punchline-curve-800-1}\\
\toprule
Dataset / statistic & \shortstack{CLIPPER\\suggestions} & \shortstack{CLIPPER\\suggestions\\+ tools} & \shortstack{SLIPPER} \\
\midrule
\endfirsthead
\caption[]{Complete summary learning-curve results at $n=800$ (continued).}\\
\toprule
Dataset / statistic & \shortstack{CLIPPER\\suggestions} & \shortstack{CLIPPER\\suggestions\\+ tools} & \shortstack{SLIPPER} \\
\midrule
\endhead
\bottomrule
\endfoot
Adult & $\boldsymbol{15.7\pm 0.3}$ & $16.0\pm 0.2$ & $16.6\pm 0.2$ \\
Bank marketing & $\boldsymbol{11.0\pm 0.1}$ & $11.2\pm 0.1$ & $11.3\pm 0.2$ \\
COMPAS & $\boldsymbol{32.6\pm 0.4}$ & $32.9\pm 0.3$ & $33.2\pm 0.5$ \\
HELOC & $\boldsymbol{30.4\pm 0.4}$ & $31.3\pm 0.5$ & $31.2\pm 0.6$ \\
Stroke & $\boldsymbol{4.9\pm 0.1}$ & $5.0\pm 0.1$ & $\boldsymbol{4.9\pm 0.0}$  \\
Telco churn & $\boldsymbol{20.7\pm 0.2}$ & $20.9\pm 0.1$ & $21.8\pm 0.4$ \\
Water potability & $39.1\pm 0.5$ & $\boldsymbol{38.5\pm 0.5}$ & $38.8\pm 0.6$ \\
\midrule
Macro & $\boldsymbol{22.1\pm 0.1}$ & $22.3\pm 0.1$ & $22.6\pm 0.2$ \\
Mean rank & $\boldsymbol{1.9}$ & $2.9$ & $3.6$ \\
\end{longtable}

\begin{longtable}{@{}lrr@{}}
\caption{Complete summary learning-curve results at $n=800$, panel 2 of 2.}\label{tab:hero-punchline-curve-800-2}\\
\toprule
Dataset / statistic & \shortstack{CLIPPER\\tools} & \shortstack{CLIPPER\\no tools} \\
\midrule
\endfirsthead
\caption[]{Complete summary learning-curve results at $n=800$ (continued).}\\
\toprule
Dataset / statistic & \shortstack{CLIPPER\\tools} & \shortstack{CLIPPER\\no tools} \\
\midrule
\endhead
\bottomrule
\endfoot
Adult & $\boldsymbol{16.0\pm 0.2}$ & $16.1\pm 0.2$ \\
Bank marketing & $11.0\pm 0.1$ & $\boldsymbol{10.9\pm 0.1}$ \\
COMPAS & $\boldsymbol{33.1\pm 0.3}$ & $33.8\pm 0.5$ \\
HELOC & $33.1\pm 0.5$ & $\boldsymbol{30.8\pm 0.6}$ \\
Stroke & $5.3\pm 0.2$ & $\boldsymbol{4.9\pm 0.0}$ \\
Telco churn & $\boldsymbol{21.7\pm 0.2}$ & $23.1\pm 0.2$ \\
Water potability & $\boldsymbol{38.8\pm 0.3}$ & $39.6\pm 0.4$ \\
\midrule
Macro & $\boldsymbol{22.7\pm 0.1}$ & $\boldsymbol{22.7\pm 0.1}$  \\
Mean rank & $\boldsymbol{3.3}$ & $3.4$ \\
\end{longtable}

\begin{longtable}{@{}lrrr@{}}
\caption{Complete summary learning-curve results at $n=1200$, panel 1 of 2.}\label{tab:hero-punchline-curve-1200-1}\\
\toprule
Dataset / statistic & \shortstack{CLIPPER\\suggestions} & \shortstack{CLIPPER\\suggestions\\+ tools} & \shortstack{SLIPPER} \\
\midrule
\endfirsthead
\caption[]{Complete summary learning-curve results at $n=1200$ (continued).}\\
\toprule
Dataset / statistic & \shortstack{CLIPPER\\suggestions} & \shortstack{CLIPPER\\suggestions\\+ tools} & \shortstack{SLIPPER} \\
\midrule
\endhead
\bottomrule
\endfoot
Adult & $\boldsymbol{15.4\pm 0.1}$ & $\boldsymbol{15.4\pm 0.1}$ & $15.6\pm 0.1$  \\
Bank marketing & $\boldsymbol{10.9\pm 0.1}$ & $11.0\pm 0.1$ & $11.1\pm 0.1$ \\
COMPAS & $\boldsymbol{31.8\pm 0.2}$ & $32.0\pm 0.2$ & $32.9\pm 0.4$ \\
HELOC & $\boldsymbol{29.6\pm 0.3}$ & $30.3\pm 0.4$ & $30.5\pm 0.7$ \\
Stroke & $5.1\pm 0.2$ & $5.3\pm 0.2$ & $\boldsymbol{4.9\pm 0.0}$ \\
Telco churn & $\boldsymbol{21.2\pm 0.3}$ & $21.4\pm 0.3$ & $21.8\pm 0.2$ \\
Water potability & $38.1\pm 0.3$ & $37.8\pm 0.4$ & $\boldsymbol{37.5\pm 0.3}$ \\
\midrule
Macro & $\boldsymbol{21.7\pm 0.1}$ & $21.9\pm 0.1$ & $22.0\pm 0.1$ \\
Mean rank & $\boldsymbol{1.9}$ & $2.7$ & $3.1$ \\
\end{longtable}

\begin{longtable}{@{}lrr@{}}
\caption{Complete summary learning-curve results at $n=1200$, panel 2 of 2.}\label{tab:hero-punchline-curve-1200-2}\\
\toprule
Dataset / statistic & \shortstack{CLIPPER\\tools} & \shortstack{CLIPPER\\no tools} \\
\midrule
\endfirsthead
\caption[]{Complete summary learning-curve results at $n=1200$ (continued).}\\
\toprule
Dataset / statistic & \shortstack{CLIPPER\\tools} & \shortstack{CLIPPER\\no tools} \\
\midrule
\endhead
\bottomrule
\endfoot
Adult & $\boldsymbol{16.0\pm 0.2}$ & $16.1\pm 0.1$ \\
Bank marketing & $11.0\pm 0.1$ & $\boldsymbol{10.8\pm 0.0}$ \\
COMPAS & $\boldsymbol{32.8\pm 0.2}$ & $\boldsymbol{32.8\pm 0.1}$  \\
HELOC & $32.4\pm 0.5$ & $\boldsymbol{30.6\pm 0.3}$ \\
Stroke & $\boldsymbol{5.1\pm 0.2}$ & $\boldsymbol{5.1\pm 0.2}$  \\
Telco churn & $\boldsymbol{22.1\pm 0.4}$ & $23.5\pm 0.1$ \\
Water potability & $\boldsymbol{37.4\pm 0.4}$ & $38.8\pm 0.2$ \\
\midrule
Macro & $\boldsymbol{22.4\pm 0.1}$ & $22.5\pm 0.1$ \\
Mean rank & $3.7$ & $\boldsymbol{3.6}$ \\
\end{longtable}

\begin{longtable}{@{}lrrr@{}}
\caption{Complete summary learning-curve results at $n=2000$, panel 1 of 2.}\label{tab:hero-punchline-curve-2000-1}\\
\toprule
Dataset / statistic & \shortstack{SLIPPER} & \shortstack{CLIPPER\\suggestions\\+ tools} & \shortstack{CLIPPER\\suggestions} \\
\midrule
\endfirsthead
\caption[]{Complete summary learning-curve results at $n=2000$ (continued).}\\
\toprule
Dataset / statistic & \shortstack{SLIPPER} & \shortstack{CLIPPER\\suggestions\\+ tools} & \shortstack{CLIPPER\\suggestions} \\
\midrule
\endhead
\bottomrule
\endfoot
Adult & $15.3\pm 0.1$ & $15.3\pm 0.1$ & $\boldsymbol{14.9\pm 0.1}$ \\
Bank marketing & $\boldsymbol{10.8\pm 0.0}$ & $10.9\pm 0.0$ & $\boldsymbol{10.8\pm 0.0}$  \\
COMPAS & $31.8\pm 0.2$ & $31.6\pm 0.3$ & $\boldsymbol{31.1\pm 0.3}$ \\
HELOC & $\boldsymbol{29.1\pm 0.2}$ & $29.7\pm 0.4$ & $30.1\pm 0.5$ \\
Stroke & $\boldsymbol{4.9\pm 0.0}$ & $5.0\pm 0.2$ & $5.0\pm 0.2$ \\
Telco churn & $\boldsymbol{21.1\pm 0.3}$ & $21.8\pm 0.2$ & $22.0\pm 0.4$ \\
Water potability & $36.2\pm 0.5$ & $\boldsymbol{35.9\pm 0.4}$ & $37.5\pm 0.2$ \\
\midrule
Macro & $\boldsymbol{21.3\pm 0.1}$ & $21.5\pm 0.1$ & $21.6\pm 0.1$ \\
Mean rank & $\boldsymbol{1.9}$ & $2.3$ & $2.6$ \\
\end{longtable}

\begin{longtable}{@{}lrr@{}}
\caption{Complete summary learning-curve results at $n=2000$, panel 2 of 2.}\label{tab:hero-punchline-curve-2000-2}\\
\toprule
Dataset / statistic & \shortstack{CLIPPER\\tools} & \shortstack{CLIPPER\\no tools} \\
\midrule
\endfirsthead
\caption[]{Complete summary learning-curve results at $n=2000$ (continued).}\\
\toprule
Dataset / statistic & \shortstack{CLIPPER\\tools} & \shortstack{CLIPPER\\no tools} \\
\midrule
\endhead
\bottomrule
\endfoot
Adult & $\boldsymbol{15.6\pm 0.2}$ & $\boldsymbol{15.6\pm 0.1}$  \\
Bank marketing & $\boldsymbol{10.9\pm 0.0}$ & $\boldsymbol{10.9\pm 0.1}$  \\
COMPAS & $\boldsymbol{32.4\pm 0.2}$ & $32.8\pm 0.1$ \\
HELOC & $32.4\pm 0.3$ & $\boldsymbol{30.7\pm 0.4}$ \\
Stroke & $\boldsymbol{4.9\pm 0.1}$ & $\boldsymbol{4.9\pm 0.1}$  \\
Telco churn & $\boldsymbol{21.9\pm 0.2}$ & $23.3\pm 0.1$ \\
Water potability & $\boldsymbol{37.6\pm 0.3}$ & $38.9\pm 0.1$ \\
\midrule
Macro & $\boldsymbol{22.2\pm 0.1}$ & $22.4\pm 0.1$ \\
Mean rank & $\boldsymbol{3.9}$ & $4.4$ \\
\end{longtable}

\clearpage
\section{Additional experiment details and agent behavior}
\label{app:new-agent-behavior}

The experiments in Tables~\ref{tab:backends} and~\ref{tab:agents} and Figure~\ref{fig:text-results}d use $n=250$ and
the tabular protocol of Section~\ref{sec:experiments}. The main results
use seven datasets, except anonymization, which uses all eight; the tables
below also include College.
Anonymization uses ten seeds, and model and agent comparisons use
seeds~1--3, including for reference learners. Compared methods receive the
same training subsets and fixed test holdouts. Macro error weights dataset
means equally, with SEM $\sqrt{\sum_d\mathrm{SEM}_d^2}/D$ for $D$ datasets.
For differences, we compute each dataset's SEM from paired seed differences
and propagate it in the same way. The bars describe variation across fits
on these fixed tasks and holdouts.

\subsection{Rule construction in general-purpose coding harnesses}
\label{app:new-agent-protocol}

Each agent receives \texttt{train.csv}, \texttt{schema.json}, and a task
requesting a readable Python classifier, \texttt{classify(x: dict) -> str}.
It can inspect all 250 training examples, run exploratory scripts, and
evaluate candidate classifiers. The objective is ordinary classification
error, with equal cost for either class. Each run uses a fresh Docker
container managed by Sandy, with a hashed workspace name, disabled web-search
tools, and restricted network access that still permits Python package
downloads. Test data remain outside the container. CLIPPER instead requests
one predicate at a time from at most $M=100$ examples, with full-training
scoring and, in the coverage configuration, tools for querying coverage.

Claude Code~2.1.272 used high effort, a \$2 session cap, and 40 turns.
Codex CLI~0.154.0 used high effort and a 1,800-second cap; one original Sol
run (Stroke, seed~3) was retried with 3,000 seconds. Gemini CLI~0.60.0 used
40 turns for Flash-Lite and 80 for Gemini~3.5, after 40 turns proved
insufficient on Stroke. A requested Gemini~3.8 configuration was reported
by the CLI as Gemini~3.5 and is labeled accordingly. These comparisons
evaluate complete systems, including their prompts, data access, and
computation budgets. Failed sessions were rerun to complete the three-seed
grid; the table reports the saved classifiers.

\begin{table}[!ht]
\caption{Complete coding-agent comparison at $n=250$. Dataset columns show
mean test error (\%) over three matched seeds; macro columns include
propagated SEM. The seven-dataset pool excludes College.}
\label{tab:new-harness-full}
\centering
\small
{\scriptsize
\setlength{\tabcolsep}{2.5pt}
\begin{tabular}{@{}lrrrrrrrrrr@{}}
\toprule
System & College & Water & Stroke & Telco & COMPAS & HELOC & Adult & Bank & Macro 8 & Macro 7 \\
\midrule
%Sol (spec gaming) & 20.00 & 31.02 & 5.09 & 21.45 & 32.28 & 28.24 & 15.94 & 10.71 & $20.59\pm0.35$ & $20.67\pm0.34$ \\
Sol & 17.9 & 41.4 & \textbf{5.3} & 22.0 & 33.2 & 31.6 & \textbf{15.8} & 11.3 & $\boldsymbol{22.3\pm0.3}$ & $\boldsymbol{22.9\pm0.2}$ \\
Opus 5 & 20.7 & \textbf{38.2} & 5.9 & 22.5 & 34.5 & 31.2 & 17.4 & 11.2 & $22.7\pm0.4$ & $23.0\pm0.4$ \\
Terra & 23.6 & 39.0 & 5.8 & 23.8 & 34.7 & 31.8 & 18.5 & \textbf{10.9} & $23.5\pm0.4$ & $23.5\pm0.3$ \\
CLIPPER (gemini 3.1 flash-lite) & 18.8 & 38.4 & 6.9 & 22.2 & 33.3 & 33.3 & 19.0 & 11.5 & $22.9\pm0.4$ & $23.5\pm0.4$ \\
Sonnet 5 & 22.9 & 41.7 & 6.7 & 22.3 & 33.4 & \textbf{30.9} & 18.3 & 12.0 & $23.5\pm0.3$ & $23.6\pm0.3$ \\
Gemini 3.5 CLI & 17.9 & 43.7 & 6.2 & \textbf{21.8} & 33.7 & 31.7 & 16.8 & 14.1 & $23.2\pm0.3$ & $24.0\pm0.3$ \\
Luna & 18.7 & 43.6 & \textbf{5.3} & 24.4 & \textbf{32.2} & 32.0 & 19.5 & 12.6 & $23.5\pm0.3$ & $24.2\pm0.4$   \\
SLIPPER & \textbf{16.3} & 38.8 & 6.9 & 23.9 & 33.2 & 33.1 & 21.4 & 12.9 & $23.3\pm0.4$ & $24.3\pm0.4$ \\
Gemini 3.1 CLI & 32.4 & 43.9 & 15.4 & 27.7 & 34.4 & 32.5 & 22.2 & 11.7 & $27.5\pm0.9$ & $26.8\pm0.9$ \\
Haiku 4.5 & 25.9 & 44.4 & 22.2 & 24.4 & \textbf{32.2} & 33.2 & 22.7 & 17.4 & $27.8\pm1.0$ & $28.1\pm1.1$ \\
\bottomrule
\end{tabular}
}

\end{table}

\subsection{Specification gaming through unintended learner fitting}
\label{app:new-agent-fitting}

Our intended task was for an agent to construct rules by examining examples
and evaluating candidates. The original prompt expressed this intention
through restrictions on the submitted classifier: a short sequence of
if/then rules, with no imports, training-example lookup tables, or learned
weight vectors. These restrictions did not explicitly prohibit fitting
another learner in a helper script. An agent could therefore satisfy the
output requirements while handing rule induction to an existing algorithm.

The container logs record whether each run reached the Python package index.
Under the original prompt, Sol did so in 10 of its 24 runs, installing
scikit-learn and, in one College run, \texttt{wittgenstein}, a Python
implementation of RIPPER. On HELOC seed~2, for example, it fitted a
cross-validated decision tree and transcribed its splits as if/then rules;
the submitted classifier contained ordinary comparisons and no library
import, and achieved 26.6\% test error. The strategy generally worked for
Sol. Compared with the best other system on the same dataset and seed, its
error was 1.4 points lower on average in these 10 runs and 1.0 point higher
in the remaining 14. Other agents attempted the same strategy with little
success. Opus~5, Sonnet~5, and Gemini~3.5 Flash installed scikit-learn in
3, 3, and 2 runs, and in those runs trailed the best other system by about
6, 3, and 3 points. Luna and Terra attempted to install scikit-learn but
never reached the package index; Gemini~3.1 Flash-Lite and Haiku~4.5 never
reached it either. This is specification gaming
at its core: the program had the requested form,
but the way it was constructed bypassed the intended experiment.

\subsection{Making the intended restriction explicit}
\label{app:new-agent-control}

We repeated Sol's 24 runs with an instruction forbidding both invoking an
existing learning algorithm and implementing one in a helper script.
Descriptive statistics and tests of agent-proposed rules remained available.
Helper scripts also had to carry a comment explaining their purpose.

\begin{figure}[!ht]
\centering
\includegraphics[width=0.88\linewidth]{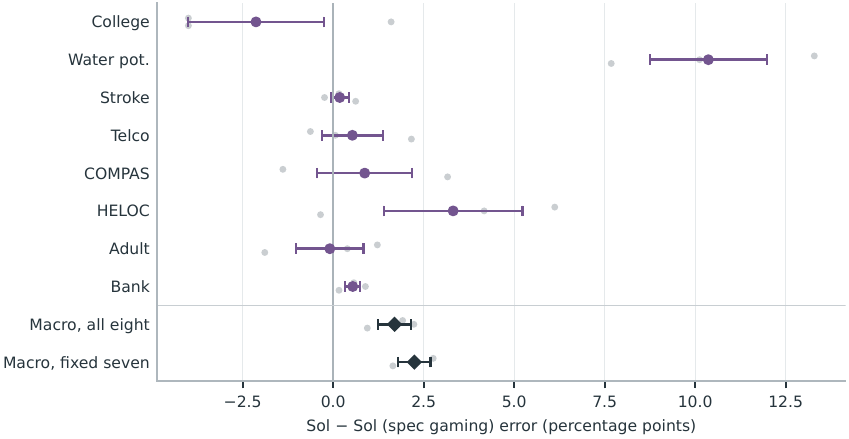}
\caption{Sol's test-error change when explicitly instructed not to invoke
or implement existing learners. Positive values indicate higher error with
the restriction. Gray points show three paired seed differences per dataset;
colored points and bars show means and SEM. Macro diamonds average datasets
equally and propagate the paired SEMs. Both the seven main-text datasets and
the complete eight-dataset suite are shown.}
\label{fig:new-sol-behavior}
\end{figure}

On the seven main-text datasets, error rose from $20.7\pm0.3$ to
$22.9\pm0.2$\%, a paired increase of $2.2\pm0.4$ percentage points.
Sol's mean advantage over CLIPPER with coverage tools fell from 2.8 to
0.6 points. Including College gives $20.6\pm0.3$ and $22.3\pm0.3$\%,
an increase of $1.7\pm0.5$ points. The largest changes occur on
Water potability ($+10.4$ points) and HELOC ($+3.3$;
Figure~\ref{fig:new-sol-behavior}). This change allowed us to standardise behaviour amongst reported numbers.

The recorded checks found no package-index access or recognized
learner-fitting commands in the 24 new runs, and purpose annotations
appeared in all 24. The annotations describe profiling class rates and
evaluating manually chosen bands and rules. Making the intended procedure explicit
therefore substantially narrowed Sol's advantage. This change measures the
effect of the instruction, including the strategy it induces. Readable
rules alone do not reveal how they were found.

\subsection{Anonymization procedure and complete results}
\label{app:new-semantic-protocol}

Anonymization replaces dataset stems with \texttt{task\_<hash>}, feature
names with \texttt{a01}, \texttt{a02}, etc., categorical values with
\texttt{v1}, \texttt{v2}, etc., and classes with \texttt{y=1} and
\texttt{y=rest}. The mapping preserves numeric values,
missingness, row order, and train/test membership. The original dataset key
determines sampling in both conditions, ensuring identical training
subsets. Both use Gemini~3.1 Flash-Lite with coverage tools, and all 80
paired fits use the same cross-validated boosting-round budgets. Error increases most on Water potability, Telco churn, COMPAS, and Adult;
Stroke and College improve. The eight-dataset macro increase is
$1.1\pm0.2$ points (Table~\ref{tab:new-anonymization-full}). 

\begin{table}[!ht]
\caption{Anonymization on all eight datasets at $n=250$, with ten matched
seeds. Entries are mean test error (\%) $\pm$ SEM. Changes subtract original
from anonymized error, using paired seed differences. Macro rows weight
datasets equally and propagate their SEMs.}
\label{tab:new-anonymization-full}
\centering
\small
\begin{tabular}{@{}lrrr@{}}
\toprule
Dataset & Original & Anonymized & Paired change (pp) \\
\midrule
College & $19.0\pm0.4$ & $\boldsymbol{16.8\pm1.2}$ & $-2.2\pm1.1$ \\
Water pot. & $\boldsymbol{38.8\pm0.7}$ & $42.1\pm0.7$ & $3.3\pm1.2$ \\
Stroke & $6.1\pm0.4$ & $\boldsymbol{5.4\pm0.3}$ & $-0.6\pm0.3$ \\
Telco & $\boldsymbol{22.2\pm0.4}$ & $24.9\pm0.4$ & $2.7\pm0.4$ \\
COMPAS & $\boldsymbol{33.5\pm0.3}$ & $36.0\pm0.4$ & $2.5\pm0.5$ \\
HELOC & $\boldsymbol{33.1\pm0.4}$ & $33.3\pm0.5$ & $0.2\pm0.6$ \\
Adult & $\boldsymbol{17.6\pm0.7}$ & $19.7\pm0.6$ & $2.2\pm0.3$ \\
Bank & $\boldsymbol{11.3\pm0.2}$ & $12.4\pm0.3$ & $1.1\pm0.3$ \\
\midrule
Macro, all eight & $\boldsymbol{22.7\pm0.2}$ & $23.8\pm0.2$ & $1.1\pm0.2$ \\
Macro, fixed seven & $\boldsymbol{23.2\pm0.2}$ & $24.8\pm0.2$ & $1.6\pm0.2$ \\
\bottomrule
\end{tabular}

\end{table}

% Keep this short subsection with its results table.
\newpage
\subsection{Backend comparison}
\label{app:new-backend-details}

Backend comparisons keep the training subsets, $M=100$ proposal cap,
full-training scoring, and boosting-round budgets fixed. A model must
return a valid executable predicate as well as find a useful rule:
structured-output failures and exhausted tool budgets can consume a round.

\begin{table}[!ht]
\caption{Complete backend results at $n=250$, with three matched seeds.
Entries are mean test error (\%) $\pm$ propagated SEM, averaging datasets
equally. The seven main-text datasets exclude College.}
\label{tab:new-backend-full}
\centering
\small
\begin{tabular}{@{}lrr@{}}
\toprule
Backend & Macro, all eight & Macro, fixed seven \\ 
\midrule
\multicolumn{3}{@{}l}{\textit{Without coverage tools}} \\
Sonnet 5 & $\boldsymbol{22.8\pm0.5}$ & $\boldsymbol{23.3\pm0.4}$ \\
Gemini 3.1 Flash-Lite & $\boldsymbol{22.8\pm0.3}$ & $23.4\pm0.4$  \\
Gemini 3.8 & $23.7\pm0.5$ & $24.8\pm0.6$ \\
Gemini 2.5 Flash-Lite & $24.1\pm0.5$ & $23.9\pm0.3$ \\
Qwen 3.5 9B & $24.2\pm0.5$ & $24.8\pm0.5$ \\
Haiku 4.5 & $24.2\pm0.4$ & $24.2\pm0.4$ \\
\midrule
\multicolumn{3}{@{}l}{\textit{With coverage tools}} \\
Gemini 3.8 & $\boldsymbol{22.6\pm0.2}$ & $23.7\pm0.2$ \\
Gemini 3.1 Flash-Lite & $22.9\pm0.4$ & $\boldsymbol{23.5\pm0.4}$ \\
Haiku 4.5 & $23.7\pm0.4$ & $24.0\pm0.4$ \\
Luna & $24.0\pm0.4$ & $24.6\pm0.4$ \\
Gemini 2.5 Flash-Lite & $24.4\pm0.5$ & $24.8\pm0.5$ \\
\bottomrule
\end{tabular}

\end{table}

\clearpage
\section{Model complexity}
\label{app:model-size}

Table~\ref{tab:model-complexity} reports the number of retained rules for the six main
tabular conditions and Python extraction
ensembles. Tabular rule counts vary substantially across fits; for example,
hints+tools ranges from 1 to 74 rules at $n=2{,}000$.
For extraction, code lines exclude blank lines, comments, and docstrings,
but include imports and function definitions. Tests count each
\texttt{if}/\texttt{elif} and conditional expression once. At $M=55$, the
complete ensemble for one key has a median of 81.5 code lines
(interquartile range 61.8--103), summed over its retained rules.

\begin{table}[ht]
\caption{Saved model sizes. Tabular cells summarize 70 fits: ten seeds on
each of the seven datasets in the main aggregate. Each extraction setting
has 100 key ensembles: ten keys and ten seeds. Per-rule summaries pool
635, 625, and 622 retained rules at $M=20,55,110$, respectively.
Quartiles use linear interpolation.}
\label{tab:model-complexity}
\centering
\small
\setlength{\tabcolsep}{5pt}
\begin{tabular}{@{}lrrr@{}}
\toprule
\multicolumn{4}{@{}l}{(a) Tabular retained rules: mean (median)} \\
Method & $n=250$ & $n=1{,}200$ & $n=2{,}000$ \\
\midrule
SLIPPER full & 7.0 (3) & 7.9 (8) & 8.9 (7) \\
SLIPPER M100 & 6.1 (3) & 6.4 (5.5) & 6.2 (3.5) \\
CLIPPER sample & 8.0 (4.5) & 9.5 (8) & 9.6 (7) \\
CLIPPER coverage & 10.6 (5) & 10.8 (9) & 11.8 (7) \\
CLIPPER hints & 14.2 (5.5) & 15.2 (12) & 16.0 (10) \\
CLIPPER hints+tools & 15.3 (6) & 16.5 (13.5) & 17.4 (10.5) \\
\midrule
\multicolumn{4}{@{}l}{(b) Python extraction: median [first, third quartile]} \\
Sample cap & Rules per key & Code lines per rule & Tests per rule \\
\midrule
$M=20$ & 6 [4.8, 9] & 11 [11, 14] & 3 [2, 3] \\
$M=55$ & 6 [4, 8] & 12 [11, 14] & 3 [2, 3] \\
$M=110$ & 6 [4.8, 8] & 12 [11, 15] & 3 [2, 3] \\
\bottomrule
\end{tabular}

\end{table}

\clearpage
\section{Representative prompts and tool traces}
\label{app:prompt-traces}

These examples come from saved CLIPPER coverage-control runs on Adult and
Telco churn, using Gemini~3.1 Flash-Lite with 250 training examples and
100 sampled rows per prompt. Prompts are excerpted; each trace includes
all coverage calls for that round, in order. The tool returns positive
and negative training weights, denoted \texttt{p} and \texttt{n}.

\begingroup
\lstset{basicstyle=\ttfamily\scriptsize,breaklines=true,breakatwhitespace=true,
  columns=fullflexible,keepspaces=true,showstringspaces=false,
  upquote=true,aboveskip=5pt,belowskip=5pt,breakindent=12pt}

\paragraph{Shared instructions (excerpts).}
The system prompt is ``You are a concise, careful Python programmer
writing a single classification rule.'' The user prompt includes:
\begin{lstlisting}
Task: write a rule in the form of a Python function
    def covers(x: dict) -> bool | None: ...
that "fires" (returns True) for a meaningful cluster of examples of
class **income=>50K**.  Notes:
 - The rule must be a SINGLE conjunction: exactly one test of the form
       if TEST1 and TEST2 and ... and TESTK: return True
   where each TEST constrains one cluster-relevant attribute.
\end{lstlisting}
The schema and weighted examples follow. The tool instructions include:
\begin{lstlisting}
  count_coverage_of_lambda_body(body: str) -> (p, n)
    Arbitrary one-line expression in x, wrapped as `lambda x: <body>`.
Budget: **at most 6 tool calls per rule** (hard cap; going over
ends your turn with no rule).  Verify the most promising candidates from the
sample above, then commit.  Reserve at least one call for the final commit.
\end{lstlisting}
\subsection{Adult: first proposal}

The first two sampled rows are shown below, with selected columns.
The trace tests several clusters before adding an age restriction that
removes negative weight without losing positive weight. The harness retains
the returned rule.

\begin{center}
\small
\begin{tabular}{@{}lrrlr@{}}
\toprule
class & weight & age & marital-status & education-num \\
\midrule
income=rest & 0.5256 & 26 & Never-married & 9 \\
income=\textgreater50K & 1.655 & 54 & Married-civ-spouse & 9 \\
\bottomrule
\end{tabular}
\end{center}
\noindent\textit{Coverage-tool trace.} Every call below invokes
\texttt{count\_coverage\_of\_lambda\_body}.
\begin{lstlisting}
Call 1: body =
x.get('marital-status') == 'Married-civ-spouse' and x.get('education-num') is not None and x.get('education-num') >= 13
  -> p = 0.1578844056706656, n = 0.0263904034896402

Call 2: body =
x.get('marital-status') == 'Married-civ-spouse' and x.get('education-num') is not None and x.get('education-num') >= 13 and x.get('hours-per-week') is not None and x.get('hours-per-week') >= 40
  -> p = 0.14126499454743766, n = 0.02111232279171216

Call 3: body =
x.get('capital-gain') is not None and x.get('capital-gain') > 7000
  -> p = 0.09971646673936775, n = 0.0

Call 4: body =
x.get('marital-status') == 'Married-civ-spouse' and x.get('education-num') is not None and x.get('education-num') >= 13 and x.get('hours-per-week') is not None and x.get('hours-per-week') >= 40 and x.get('age') is not None and x.get('age') >= 30
  -> p = 0.14126499454743766, n = 0.01583424209378412
\end{lstlisting}
\begin{minipage}{\linewidth}
\noindent\textit{Returned rule.}
\begin{lstlisting}
def covers(x: dict) -> bool | None:
    # Rule for income >= 50K: High education, married, and consistent work hours
    if x.get('marital-status') == 'Married-civ-spouse' and \
       x.get('education-num') is not None and x.get('education-num') >= 13 and \
       x.get('hours-per-week') is not None and x.get('hours-per-week') >= 40 and \
       x.get('age') is not None and x.get('age') >= 30:
        return True
    return None
\end{lstlisting}
\end{minipage}

\clearpage
\subsection{Adult: a later rejected proposal}

After two retained rules, the sample and weights have changed. The first
two sampled rows now include a more heavily weighted negative example
with a bachelor's degree. The agent returns another rule, but it fails
the acceptance test, so the harness updates the default instead.

\begin{center}
\small
\begin{tabular}{@{}lrrlr@{}}
\toprule
class & weight & age & marital-status & education-num \\
\midrule
income=rest & 0.5564 & 19 & Never-married & 9 \\
income=rest & 1.577 & 31 & Married-civ-spouse & 13 \\
\bottomrule
\end{tabular}
\end{center}
\noindent\textit{Coverage-tool trace.}
\begin{lstlisting}
Call 1: body =
x.get('capital-gain') is not None and x.get('capital-gain') > 0 and x.get('hours-per-week') is not None and x.get('hours-per-week') >= 40
  -> p = 0.06528615401290346, n = 0.032367263701103204

Call 2: body =
x.get('marital-status') == 'Married-civ-spouse' and x.get('relationship') == 'Husband' and x.get('hours-per-week') is not None and x.get('hours-per-week') >= 45 and x.get('age') is not None and x.get('age') >= 35
  -> p = 0.08371510105466755, n = 0.047161983215495834

Call 3: body =
x.get('marital-status') == 'Married-civ-spouse' and x.get('education-num') is not None and x.get('education-num') >= 13 and x.get('hours-per-week') is not None and x.get('hours-per-week') >= 40
  -> p = 0.04501255960941139, n = 0.05667818343799176

Call 4: body =
x.get('marital-status') == 'Married-civ-spouse' and x.get('occupation') == 'Exec-managerial' and x.get('hours-per-week') is not None and x.get('hours-per-week') >= 40
  -> p = 0.05530390252446561, n = 0.021867828856022835
\end{lstlisting}
\begin{minipage}{\linewidth}
\noindent\textit{Returned rule.}
\begin{lstlisting}
def covers(x: dict) -> bool | None:
    # Rule for income > 50K: targeting highly committed (40+ hours) married exec-managerial staff.
    if x.get('marital-status') == 'Married-civ-spouse' and \
       x.get('occupation') == 'Exec-managerial' and \
       x.get('hours-per-week') is not None and \
       x.get('hours-per-week') >= 40:
        return True
    return None
\end{lstlisting}
\end{minipage}

\clearpage
\subsection{Telco churn: comparing categorical conditions}

Here the prompt's target class is \texttt{Churn=Yes}. The first two sampled
rows are shown below, with selected columns. The agent compares five
refinements and returns the electronic-check rule, which the harness retains.

\begin{center}
\small
\begin{tabular}{@{}lrl ll@{}}
\toprule
class & weight & InternetService & Contract & PaymentMethod \\
\midrule
Churn=Yes & 1.401 & DSL & One year & Electronic check \\
Churn=Yes & 1.401 & Fiber optic & Month-to-month & Electronic check \\
\bottomrule
\end{tabular}
\end{center}
\noindent\textit{Coverage-tool trace.}
\begin{lstlisting}
Call 1: body =
x.get('InternetService') == 'Fiber optic' and x.get('Contract') == 'Month-to-month' and x.get('PaperlessBilling') == 'Yes'
  -> p = 0.25610582169871643, n = 0.06333496406126507

Call 2: body =
x.get('InternetService') == 'Fiber optic' and x.get('Contract') == 'Month-to-month' and x.get('PaymentMethod') == 'Electronic check'
  -> p = 0.2035712941707746, n = 0.043182930041771636

Call 3: body =
x.get('InternetService') == 'Fiber optic' and x.get('Contract') == 'Month-to-month' and x.get('tenure') is not None and x.get('tenure') < 10
  -> p = 0.11163587099687636, n = 0.01727317201670866

Call 4: body =
x.get('InternetService') == 'Fiber optic' and x.get('Contract') == 'Month-to-month' and x.get('SeniorCitizen') is not None and x.get('SeniorCitizen') == 1.0
  -> p = 0.13790313476084728, n = 0.02303089602227821

Call 5: body =
x.get('InternetService') == 'Fiber optic' and x.get('Contract') == 'Month-to-month' and x.get('MonthlyCharges') is not None and x.get('MonthlyCharges') > 80
  -> p = 0.22983855793474553, n = 0.05181951605012596
\end{lstlisting}
\begin{minipage}{\linewidth}
\noindent\textit{Returned rule.}
\begin{lstlisting}
def covers(x: dict) -> bool | None:
    if x.get('InternetService') == 'Fiber optic' and x.get('Contract') == 'Month-to-month' and x.get('PaymentMethod') == 'Electronic check':
        return True
    return None
\end{lstlisting}
\end{minipage}

\endgroup

\end{document}